%% file: root.tex
\documentclass{article}
\usepackage{amssymb}

\usepackage{xcolor}
\definecolor{orangelink}{RGB}{255,127,0}

\PassOptionsToPackage{numbers,square,sort&compress}{natbib}
\PassOptionsToPackage{colorlinks=true, linkcolor=black, citecolor=black,
}{hyperref}
\usepackage{lipsum}

\usepackage[preprint]{style/corl_2025} %

\usepackage{multicol}
\usepackage{amsmath, amsfonts, amssymb, amsthm}
\usepackage{enumerate}
\usepackage[inline]{enumitem}
\usepackage{mathtools}
\usepackage{comment}
\usepackage{graphicx}
\usepackage{longtable,tabularx}
\usepackage{placeins} 
\usepackage{float}
\usepackage{multirow}
\usepackage{bbm}
\usepackage{threeparttable}
\usepackage{balance}
\usepackage[ruled,algo2e]{algorithm2e}
\usepackage{algpseudocode}
\usepackage{adjustbox}
\usepackage{booktabs}
\usepackage{bm}
\usepackage{etoolbox}
\usepackage{microtype}
\usepackage[title]{appendix}
\usepackage{units}
\usepackage{caption}
\usepackage{cleveref}
\usepackage{graphicx, float, url, wrapfig, stfloats}
\usepackage{cancel}

\newcommand*{\subj}{\ensuremath{\mathrm{subject\ to\ }}}

\newcommand{\mcal}[1]{\mathcal{#1}}
\newcommand{\mbb}[1]{\mathbb{#1}}

\newcommand{\Expect}{\ensuremath{\mathbb{E}}}

\newcommand{\norm}[1]{\left\Vert #1 \right\Vert}

\newcommand{\algname}{\mbox{WoMAP}\xspace}

\newbool{extended}
\setbool{extended}{false}

\newbool{rss_format}
\setbool{rss_format}{false}

\makeatletter
\newcommand{\longdash}[1][2em]{%
  \makebox[#1]{$\m@th\smash-\mkern-7mu\cleaders\hbox{$\mkern-2mu\smash-\mkern-2mu$}\hfill\mkern-7mu\smash-$}}
\makeatother
\newcommand{\omitskip}{\kern-\arraycolsep}

\title{WoMAP: World Models For Embodied Open-Vocabulary Object Localization}

\author{
 Tenny Yin$^{1}$\thanks{Equal contribution. $^{\dagger}$Authors contributed equally.}\quad 
 Zhiting Mei$^{1}$\quad
 Tao Sun$^{1,2}$\quad \\
 \textbf{Lihan Zha}$^{1}$\quad
 \textbf{Jeremy Bao}$^{1\dagger}$\quad
 \textbf{Miyu Yamane}$^{1\dagger}$\quad \\
  \textbf{Emily Zhou}$^{1\dagger}$\quad
 \textbf{Ola Shorinwa}$^{1*}$\quad
 \textbf{Anirudha Majumdar}$^{1}$\quad\\
 $^{1}$Princeton University \quad
 $^{2}$McGill University \\
  \small{\href{https://robot-womap.github.io}{{\color{orange} \textbf{robot-womap.github.io}}}}
  \vspace{-1.5em}
}

\begin{document}
    \makeatletter
    \let\@oldmaketitle\@maketitle %
    \renewcommand{\@maketitle}{
        \@oldmaketitle%
    }
    \makeatother
    
\maketitle

\input{sections/abstract}

\keywords{
Active Perception, World Models, Object Localization
}

\begin{center}
    \captionsetup{type=figure} 
    \includegraphics[width=1.0\linewidth]{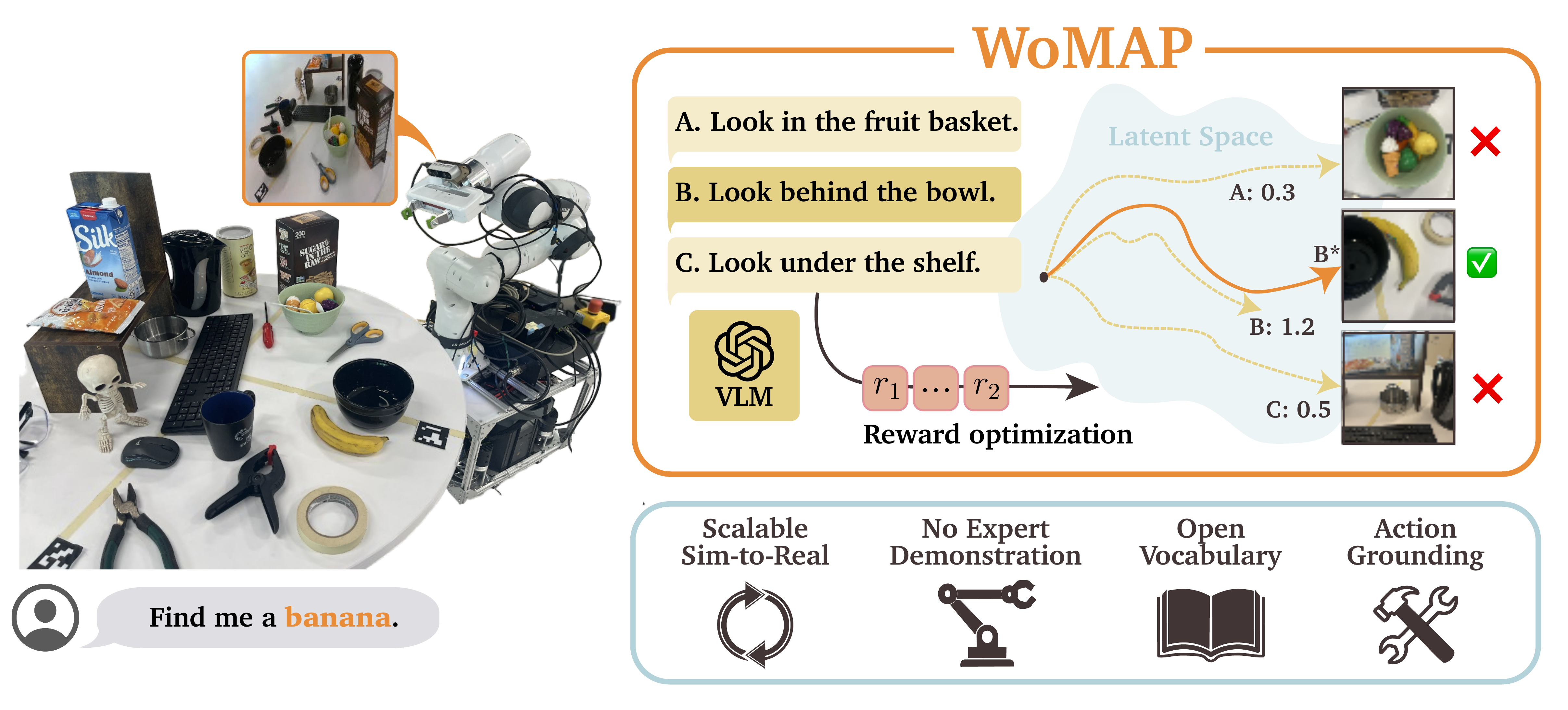}
    \captionof{figure}{\textbf{WoMAP} uses a world model to ground high-level action proposals and maximize predicted rewards. In this example, given three high-level VLM proposals, WoMAP selects ``look behind the bowl'' as the optimal choice after evaluating outcomes of each action roll-out in latent space.}
    \label{fig:womap_banner}
\end{center}

\input{sections/1-introduction}

\input{sections/2-related_work}

\input{sections/3-womap}

\input{sections/experiments}

\input{sections/conclusion}

\input{sections/limitations_future_work}

\acknowledgments{
The authors were partially supported by the NSF CAREER Award \#2044149, the Office of Naval Research (N00014-23-1-2148), a Sloan Fellowship, and the Princeton SEAS Innovation Fund. 

}

\bibliography{references.bib}

\clearpage

\begin{appendices}
    \input{sections/final_appendix}

\end{appendices}

\end{document}

%% file: sections/abstract.tex
\begin{abstract}
Language-instructed active object localization is a critical challenge for robots, requiring efficient exploration of partially observable environments.
However, state-of-the-art approaches either struggle to generalize beyond demonstration datasets (e.g., imitation learning methods) or fail to generate physically grounded actions (e.g., VLMs).
To address these limitations, 
we introduce \textbf{WoMAP} (\textbf{Wo}rld \textbf{M}odels for \textbf{A}ctive \textbf{P}erception): a recipe for training open-vocabulary object localization policies that: (i) uses a Gaussian Splatting-based real-to-sim-to-real pipeline for scalable data generation without the need for expert demonstrations, (ii) distills dense rewards signals from open-vocabulary object detectors, 
and (iii) leverages a latent world model for dynamics and rewards prediction to ground high-level action proposals at inference time.
Rigorous simulation and hardware experiments demonstrate WoMAP's superior performance in a broad range of zero-shot object localization tasks, with more than $9\times$ and $2\times$ higher success rates compared to VLM and diffusion policy baselines, respectively. 
Further, we show that WoMAP achieves strong generalization and sim-to-real transfer on a TidyBot. %

\end{abstract}

%% file: sections/1-introduction.tex
\section{Introduction}
\label{sec:introduction}
Perceptual activity in biological agents is inherently active and exploratory \cite{plasticq, 5968}. As an example, consider the task of \emph{open-vocabulary object localization}, where an agent needs to approach a target object specified by natural language in a previously unseen environment. 
In such settings, humans will actively seek information guided by prior expectations to search efficiently. For example, when looking for keys, we preferentially inspect locations where they are most likely to be found, e.g., near the door or on the couch.

However, reproducing intelligent search behavior for robots remains challenging, as it requires semantic understanding of open-vocabulary object descriptions and commonsense reasoning from partial observations in unfamiliar environments.
While vision-language models (VLMs) provide useful heuristics for exploration~\cite{chang2023goatthing, chaplot2020object, wen2025zero, ren2024explore}, effectively grounding these high-level action proposals to physical execution is a non-trivial problem. This grounding can be achieved via imitation learning methods~\cite{ramrakhya2023pirlnav, yokoyama2024hm3d}, which require large-scale expert demonstrations and can struggle to generalize beyond demonstration datasets. Alternatively, reinforcement learning (RL) \cite{ye2021auxiliary, fan2023evidentialactiverecognitionintelligent, Ramakrishnan_2019} offers another route to grounding, but is challenging to employ without an accurate simulation environment.

To address these challenges, we present \textbf{WoMAP} (\textbf{Wo}rld \textbf{M}odels for \textbf{A}ctive \textbf{P}erception): a novel recipe for efficient active object localization that can be trained without expert demonstrations or online interactions with the environment (\Cref{fig:womap_banner}). 
In order to achieve this, we propose an approach that learns a \emph{latent world model}~\cite{ha2018world} using three key ingredients (\Cref{fig:pipeline}, left). First, we introduce a scalable data generation pipeline based on Gaussian Splatting~\cite{kerbl20233d} that allows us to generate photorealistic data with broad coverage using real-world videos. Second, we propose a training framework that is \emph{reconstruction-free}; instead of using image reconstruction as a supervisory training signal (which can lead to poor generalization, training stability, and sample efficiency~\cite{zhou2024dino}), we construct dense rewards from the confidence outputs of open-vocabulary object detectors and \emph{distill} these into the latent space of the world model. 
Finally, we present an inference-time planning scheme that optimizes high-level action proposals from VLMs using the trained world model.

Taken together, we contribute a novel approach to open-vocabulary object localization that can be trained in a data-efficient manner, generalize to novel scenes and object descriptions, and exploit commonsense reasoning abilities of VLMs. 
We demonstrate our approach on a suite of simulated and real-world object localization tasks with significant improvements ($2\times$ -- $9\times$ higher success rates) over baselines that only utilize imitation learning or VLMs.

%% file: sections/2-related_work.tex
\section{Related Work}
\label{sec:related_work}
\noindent\textbf{Active Object Localization.} 
Broadly, active object localization has been explored with both end-to-end approaches \cite{ramrakhya2022habitat, gervet2023navigating}, such as imitation learning (IL) \cite{ramrakhya2023pirlnav, yokoyama2024hm3d} and reinforcement learning (RL) \cite{ye2021auxiliary, fan2023evidentialactiverecognitionintelligent, Ramakrishnan_2019}, and modular approaches incorporating foundation models \cite{chang2023goatthing, chaplot2020object, wen2025zero}.
End-to-end methods map visual observations directly to actions but typically require large amounts of expert demonstrations or interactions to learn effective exploration behaviors \cite{ramrakhya2022habitat, ramrakhya2023pirlnav} and generalize poorly to new environments or tasks \cite{zhou2023esc}. 
In contrast, WoMAP does not require on-policy demonstrations or interactions with the environment, and leverages a latent world model to plan sequences of actions at inference time in order to achieve strong generalization across tasks and sim-to-real gaps.

Modular approaches incorporate foundation models such as pre-trained object detectors or VLMs \cite{chaplot2020object, wen2025zero, jiang2024roboexp} to reason over observations and plan exploration.
However, they rely heavily on the accuracy of each system component and require engineered spatial scene representations to ground executions in the real world.
Instead, WoMAP directly optimizes VLM outputs within a learned environment model, offering a light-weight solution for grounded actions with minimal reliance on external representations.
Finally, in terms of task setup, most prior work focuses on simulated indoor navigation, utilizing rich contextual cues (e.g., sofas are more likely to be in the living room) and restricting action spaces for tractability. In contrast, our framework makes no such assumptions and tackles more general settings requiring full 6-DoF camera control to locate objects in cluttered scenes.

\noindent\textbf{World Models for Robotics.}
World models have become increasingly prominent in robotics, providing predictive foresight in learning action-conditioned dynamics for long-horizon planning  \cite{mendonca2023alan, wu2023daydreamer, zhou2024dino, nakamura2025generalizing}. 
To capture environment dynamics that generalize to test-time distributions, existing works typically rely on large quantities of uncurated data \cite{wu2023daydreamer, zhou2024dino}, expert demonstrations \cite{nakamura2025generalizing, bar2024navigation}, or on-policy interactions \cite{hansen2023td}, which are labor-intensive and costly to collect in real-world settings. In contrast, WoMAP learns a policy-agnostic environment model from synthetically generated offline data via Gaussian Splatting~\cite{kerbl20233d}. 
Many existing works also employ an image reconstruction loss to provide dense learning signals \cite{wu2023daydreamer,qi2025strengthening}, which often results in greater model complexity and unstable training. Without image reconstruction, WoMAP instead leverages dense reward signals distilled from pretrained models to encode rich visual, spatial, and semantic information in a pretrained latent representation, 
improving both scalability and robustness for downstream tasks.

%% file: sections/3-womap.tex
\section{Method}
\label{sec:womap}
\subsection{Problem Formulation}
We consider an open-vocabulary object localization task with a robot equipped with an onboard RGB camera with a six degree-of-freedom (6-DoF) action space, operating in an environment $E \in \mcal{E}$.
We model the problem as a partially observable Markov decision process (POMDP), where given the current state of the environment and the robot, the camera returns a partial observation $o_t$ under sensing uncertainty, occlusions, and limited field of view. 
At each time step, the robot selects a continuous action $a_t \in \mathbb{R}^6$, corresponding to camera translation and rotation in 3D space, to obtain a new observation. 
Given a language description $l$ of the target object $\mcal{T}_{g}$, the robot seeks to efficiently obtain an \emph{optimal} view (i.e., a view that maximizes the object's visibility for localization) over a planning horizon $T$: $\max_{a_{0:T}} \mathcal{R}(o_T, \mcal{T}_{g})$, where $\mathcal R \in [0, 1]$ is the object localization reward describing how well the target object $\mcal{T}_{g}$ is identified in $o_T$. 

\begin{figure*}[t]
    \centering\includegraphics[width=\linewidth]{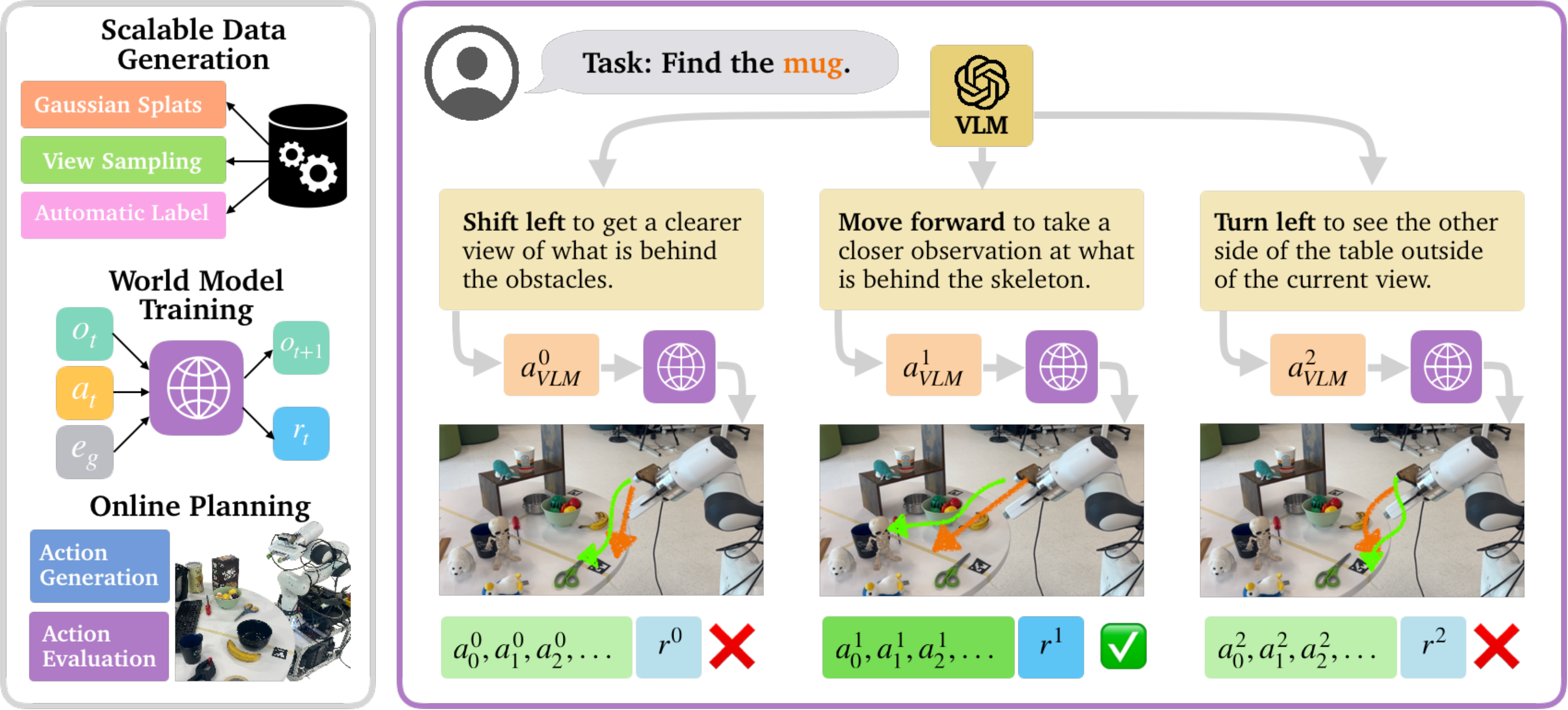}
    \caption{\textbf{Left: Core components of WoMAP.}
    Scalable data generation with Gaussian Splats (Section~\ref{subsec:womap_data_generation}), world modeling with object detection reward supervision (Section~\ref{subsec:womap_world_model}), and latent space action planning (Section~\ref{subsec:womap_planning}). \textbf{Right: The action optimization process.} 
    Given the task and current observation, a VLM generates high-level proposals, which we transform into coarse actions (green arrows); each action is further optimized within WoMAP’s reward gradient field (red arrows), and the action sequence with the highest predicted reward is executed.
    }
    \label{fig:pipeline}
\end{figure*}

Our proposed framework, WoMAP, uses a world model to capture latent space dynamics and reward predictions that can generalize to any $E \in \mcal{E}$. However, as discussed in \Cref{sec:introduction}, learning a world model that operates across such diverse task settings is non-trivial, requiring training data with sufficient coverage, strong supervisory reward signals, and the ability to incorporate high-level commonsense reasoning during planning.
In the following sections, we describe the three core components of WoMAP as illustrated in \Cref{fig:pipeline}, addressing each of these fundamental challenges.

\subsection{Scalable Data Generation}
\label{subsec:womap_data_generation}
Unlike imitation learning methods, world models do not require expert trajectories for training, which are generally expensive to collect. However, they require sufficient data coverage to effectively capture the dynamics of the environment, which is challenging to scale as the number of training environments increases. Gathering diverse observation data from the real-world also poses significant challenges, which necessitates strategic data collection to maximize sample efficiency. 

\ifbool{rss_format}
{
\begin{figure}
    \centering
    \includegraphics[width=\linewidth]{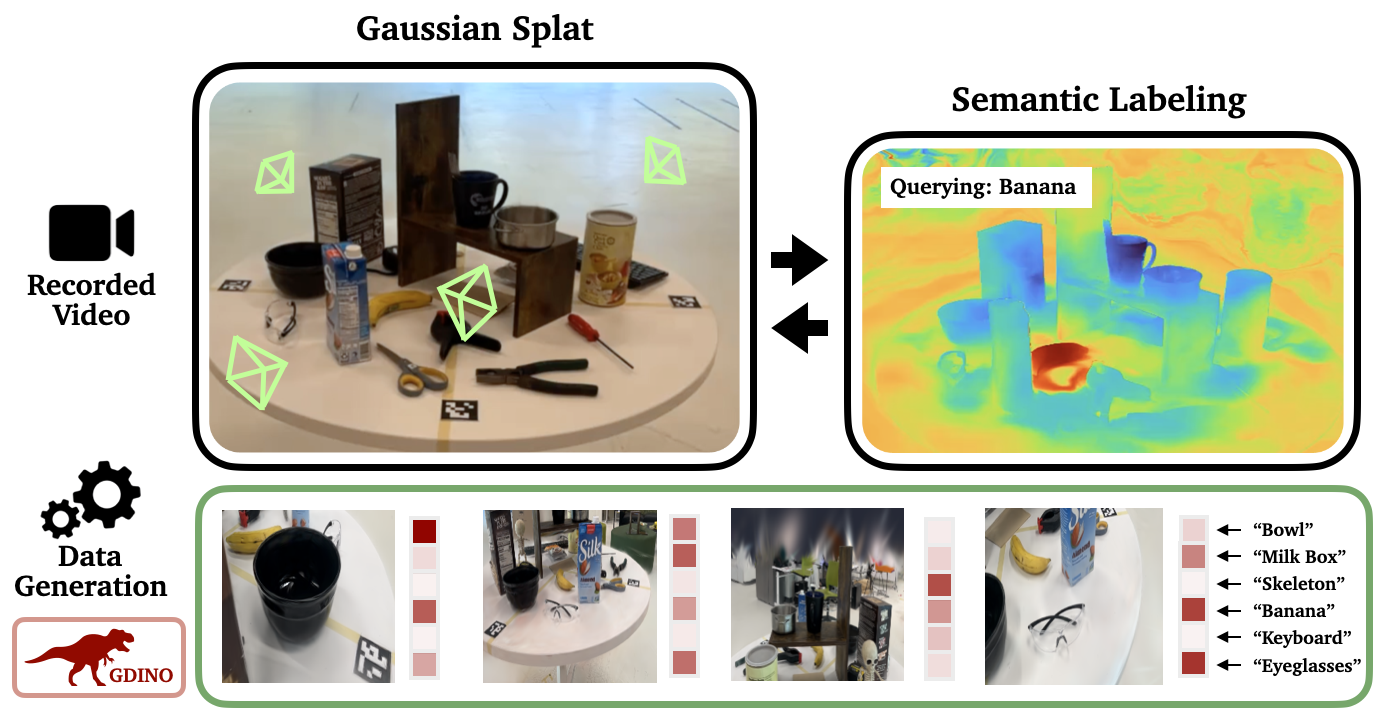}
    \caption{\textbf{Data Generation with Gaussian Splats.} We train Gaussian Splats for each scene and obtain ground truth object locations through semantic labeling \cite{shorinwa2025siren} for informative view sampling. Each observation is labeled with GroundingDINO \cite{liu2024grounding} to get confidence scores for all training targets. }
    \label{fig:gaussian-datagen}
\end{figure}
}
{
\begin{wrapfigure}{r}{0.5\linewidth}
    \vspace{-12pt}
    \centering
    \includegraphics[width=\linewidth]{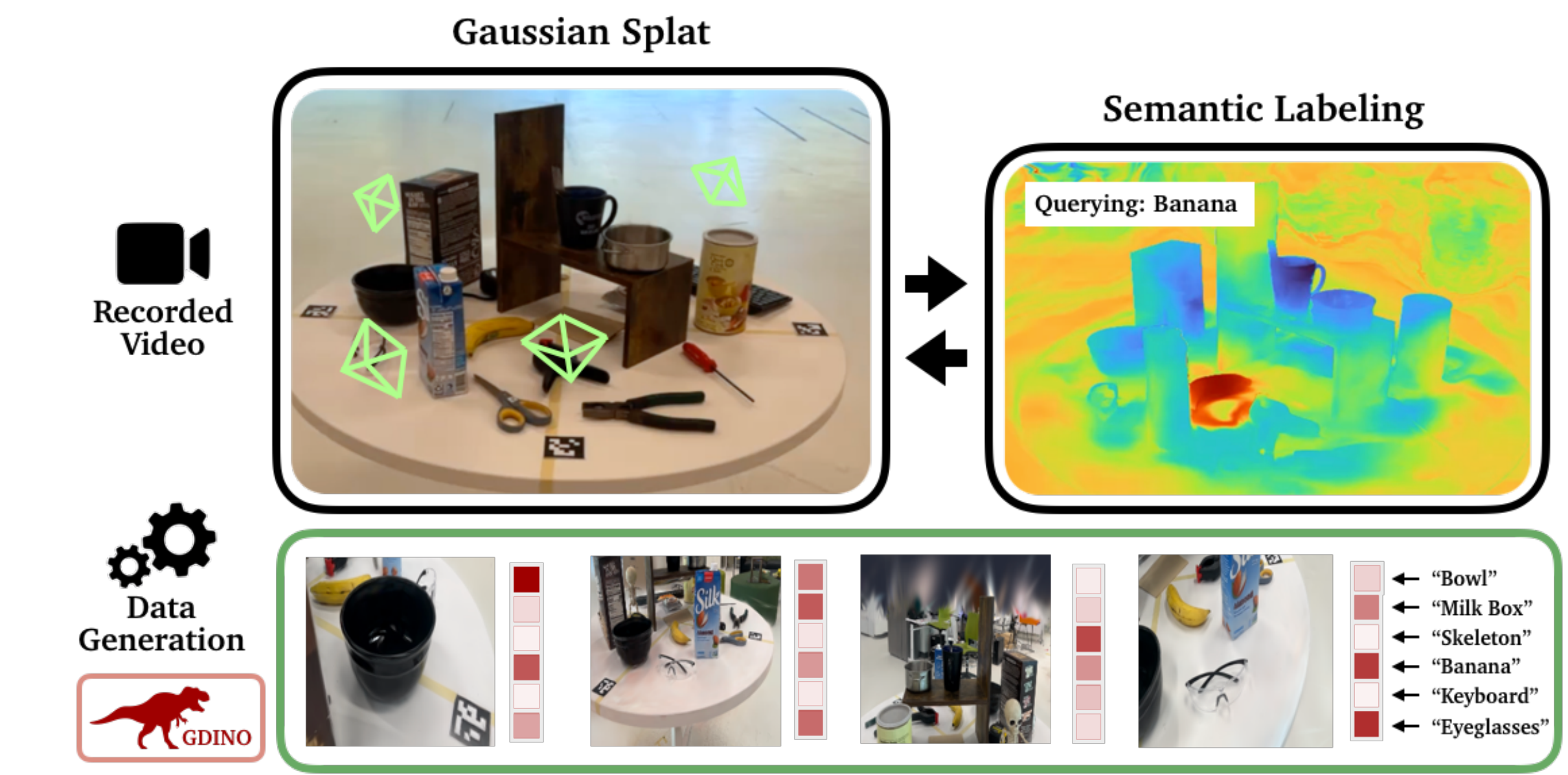}
    \caption{\textbf{Data Generation with Gaussian Splats.} We train Gaussian Splats for each scene and obtain ground truth object locations through semantic labeling \cite{shorinwa2025siren} for informative view sampling. Each observation is labeled with GroundingDINO \cite{liu2024grounding} to get confidence scores for all training targets. }
    \vspace{-5pt}
    \label{fig:gaussian-datagen}
\end{wrapfigure}
}

In WoMAP, we introduce a scalable real-to-sim-to-real data generation pipeline that utilizes only a few real-world videos to efficiently generate diverse training data.
Our pipeline leverages Gaussian Splatting \cite{kerbl20233d} to generate a photorealistic simulation environment from video input and can render arbitrary views at any camera pose. 
Shown in~\Cref{fig:gaussian-datagen}, we automatically annotate the location and physical dimension of each target in the training scene by distilling language semantics from CLIP~\cite{radford2021learning} into semantic Gaussian Splats \cite{shorinwa2025siren}. 
Within the trained Gaussian Splat, we collect a training dataset $\mathcal{D}$ consisting of $M$ observation-reward-pose tuples, i.e., %
${\mcal{D} = \{(o_{i}, r_{i}, P_{i}), \ \forall i \in [M] \}}$, %
where $P_{i} \in \mbb{R}^{6}$ denotes the spatial camera pose, $o_i$ is the associated rendered image, and $r_{i}$ is the reward associated with $o_{i}$.
At training time, given two random samples, we compute the action $a_{ij}$ required to transition from $P_{i}$ to $P_j$, since training the world model does not require sequentially-ordered data. 

Further, to improve sample efficiency, we design our data distribution to concentrate on trajectories starting from random initial positions and leading towards sampled target objects, with added linear and angular perturbations for data augmentation. These trajectories can be generated with any planning algorithm and require no human demonstration. We provide more implementation details in Appendix~\ref{appendix-impl-data-gen}.
In Section~\ref{sec:experiments_sim_to_real_transfer}, we demonstrate that despite only training on synthetically rendered images in Gaussian Splats, WoMAP still achieves strong zero-shot sim-to-real performance.

\ifbool{rss_format}
{
\begin{figure}
    \centering
    \includegraphics[width=\linewidth]{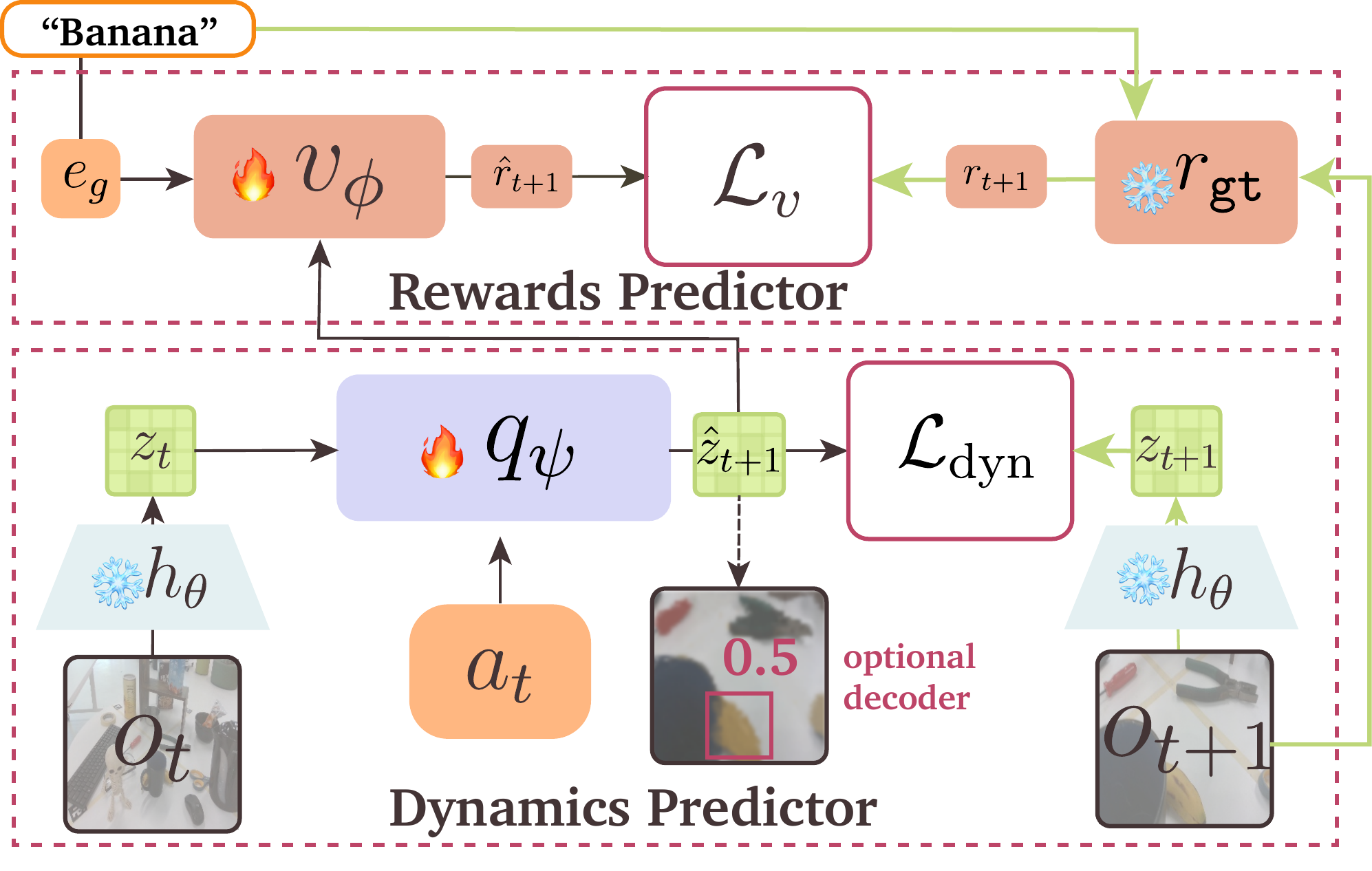}
    \caption{\textbf{World Model Architecture} for simultaneous dynamics and rewards prediction.}
    \label{fig:world_model}
\end{figure}
}
{
\begin{wrapfigure}[8]{R}{0.5\linewidth}
    \centering
    \vspace{-40pt}
    \includegraphics[width=0.9\linewidth]{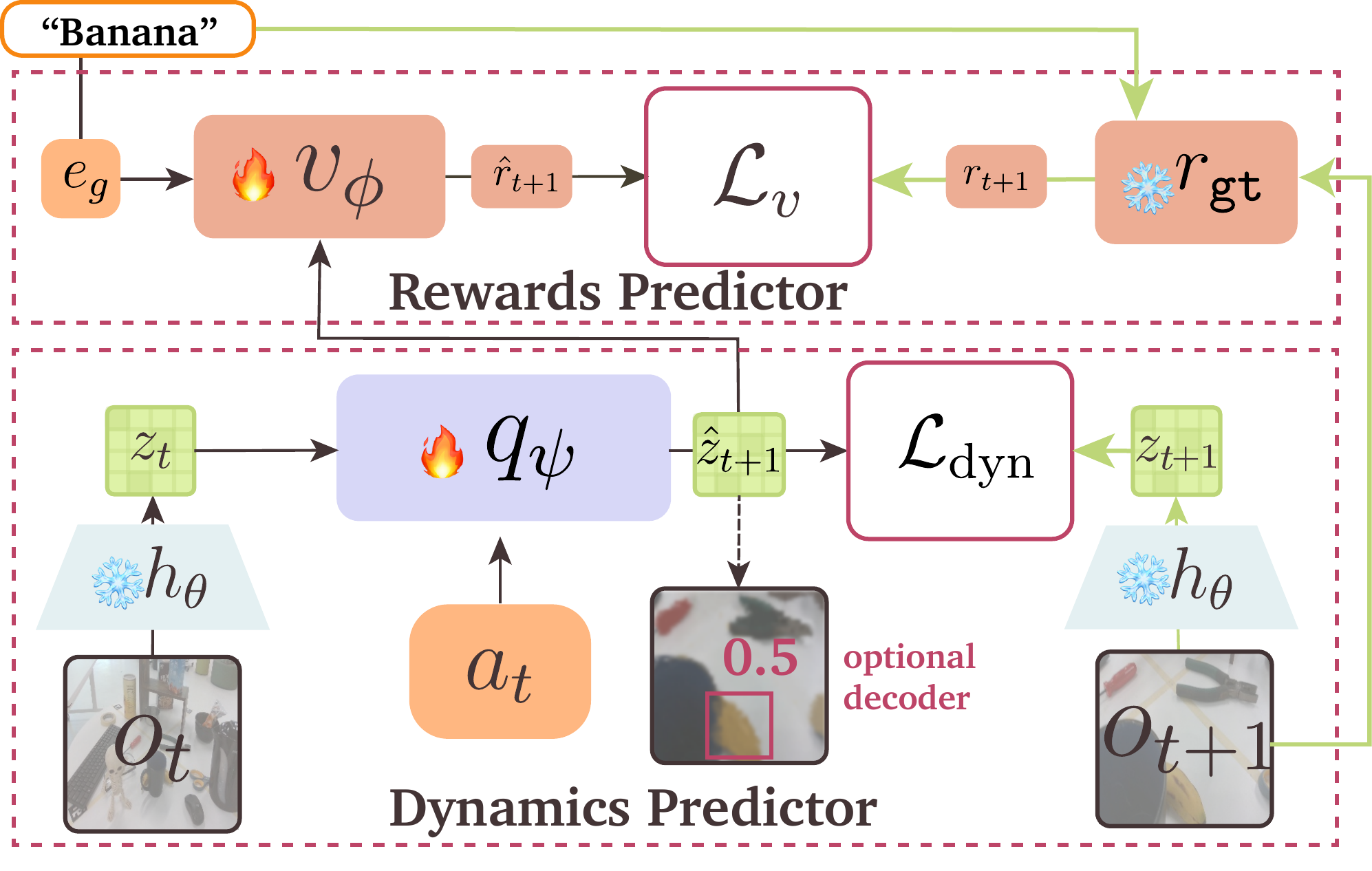}
    \caption{\textbf{World Model Architecture} for simultaneous dynamics and rewards prediction. 
    }
    \vspace{-40pt}
    \label{fig:world_model}
\end{wrapfigure}
}

\subsection{World Models for Active Perception}
\label{subsec:womap_world_model}
Given data generated from Sec.~\ref{subsec:womap_data_generation}, we outline key design choices that enable the world model to accurately learn dynamics and rewards. A central innovation of our pipeline is the use of dense reward distillation from open-vocabulary object detectors, which allows for data-efficient training without relying on image reconstruction objectives.

\subsubsection{World Model Architecture}

As shown in \Cref{fig:world_model}, the world model consists of three standard core components~\cite{ha2018world, hafner2019dream}:
\begin{equation}
    \begin{aligned}
        &\text{Observation Encoder:} && z_{t} = h_{\theta}(o_{t}), \\
        &\text{Dynamics Predictor:} && z_{t + 1} \sim q_{\psi}(z_{t + 1} \mid z_{t}, a_{t}), \\
        &\text{Rewards Predictor:} && r_{t} \sim v_{\phi}(r_{t} \mid z_{t}, e_{g}),
    \end{aligned}
\end{equation}
where ${z_{t} \in \mbb{R}^{d}}$ denotes the latent state, $e_g$ represents the language embedding computed from a description $l$ of the target object $\mcal{T}_{g}$, ${r_{t} \in \mbb{R}}$  denotes the associated reward with $z_t$ when querying for $ \mcal{T}_{g}$, and $\theta, \psi, \phi$ denote network parameters for each component of the world model.

The observation encoder $h_\theta$ maps a high-dimensional camera observation $o_t$ to a compact latent space $z_t$. 
Inspired by \cite{zhou2024dino}, we leverage a pre-trained vision encoder model DINOv2 \cite{oquab2023dinov2} to directly compute flattened patch embeddings of the given image as $z_t$ to retain rich visual and spatial features resulting from large-scale pre-training. 
The dynamics predictor $q_{\psi}$ models the transition distribution $p(z_{t + 1} \mid z_{t}, a_{t})$ using a standard ViT architecture~\cite{dosovitskiy2020image}. 
With variational inference, we parametrize $q_{\psi}$ as a Gaussian distribution ${q_{\psi}(z_{t + 1} \mid z_{t}, a_{t})} \sim \mathcal{N}(\mu_t, \sigma_t^2)$ and minimize the Kullback-Leibler (KL) divergence between the true state transition, $p(z_{t + 1} \mid z_{t}, a_{t})$ and $q_{\psi}$, with the loss function: ${\mcal{L}_{\text{dyn}} = \mathrm{KL}\big(p(z_{t + 1} \mid z_{t}, a_{t}\big) \Vert q_{\psi}(z_{t + 1} \mid z_{t}, a_{t}))}$. During training, we supervise $q_{\psi}$ recurrently on a sequence of $H$ observation-action pairs $\{(o_i, a_i)\}_{i=1}^H$ to enforce dynamics consistency.
The rewards predictor $v_{\phi}$ estimates the reward for each latent state, conditioned on the language embedding of the task $\mcal{T}_{g}$.
Additional implementation details can be found in Appendix~\ref{ssec:appendix_world_model_implementation}.

\subsubsection{Reward Distillation}
Despite providing dense reward signals, image reconstruction objectives often lead to training instability \cite{hansen2022temporal, burchi2024mudreamer}, which we further demonstrate in Appendix~\ref{appendix:ablation-encoder}.
To tackle this challenge, WoMAP introduces a novel reward distillation procedure that generates dense rewards signals without reconstruction objectives for data-efficient training. As shown in Figure~\ref{fig:gaussian-datagen}, during data generation, WoMAP computes a per-frame reward
for each object in the observation
using the detection confidence provided by a pretrained object detector, e.g., GroundingDINO \cite{liu2024grounding}, scaled by the associated detection bounding-box size. 
This annotation procedure yields a rich, task-relevant training signal for each object in the scene, and scales efficiently with environment complexity by enabling parallel processing of detections.
At training time, we distill the ground-truth reward signal $r_{\texttt{gt}}(o_{t+1}, e_{g}) \in [0, 1]$ 
into the rewards predictor $v_\phi$ conditioned on the language embedding $e$ of each relevant object.
This distillation pipeline enables an effective planning framework using world models, which we discuss in the following subsection.

\subsection{Planning with WoMAP}
\label{subsec:womap_planning}

While world models can directly plan actions via sampling or gradient-based optimization, such methods are often inefficient in continuous action spaces without informed guidance. 
Particularly, in complex problem spaces, gradient-based methods struggle to localize target objects within a finite optimization budget and frequently converge to suboptimal solutions.
WoMAP addresses this limitation by leveraging VLMs' commonsense reasoning to generate informed action proposals, which are then refined via model predictive control using a world model for spatial grounding.

For a given task instruction, we prompt a VLM using chain-of-thought prompting \cite{wei2022chain} to provide high-level guidance for promising locations for the robot to explore. In preliminary experiments, we observed that VLMs struggle with spatial understanding when prompted for numerical relative actions given an input image. Consequently, we query the VLM using a multiple-choice prompt with the options given by a fixed set of textual description of the possible actions, e.g., ``turn left," or ``move forward." We provide additional implementation details in Appendix~\ref{ssec:appendix_womap_planning}. Subsequently, WoMAP optimizes a set of candidate actions provided by the VLM to maximize the expected rewards:
\begin{equation}
    \label{eq:womap_mpc}
    \begin{aligned}
        &\max_{a_{t:t + T}} \quad \sum_{\tau = 1}^{T} (\Expect_{v_{\phi}}[r_{t + \tau} \mid z_{t + \tau}, e_{g})] + \gamma \norm{a_{t + \tau - 1} - a_{t + \tau - 2}}_{1}), \\
        &\subj \quad  z_{t + \tau} \sim q_{\psi}(z_{t + \tau} \mid z_{t + \tau - 1}, a_{t + \tau - 1}) \enspace  \forall \tau \in [T],
    \end{aligned}
\end{equation}
at each timestep $t$ with latent state $z_{t}$,  target-object language embedding $e_g$, previous control action $a_{t - 1}$, MPC planning horizon $T$, and weight ${\gamma \in \mbb{R}_{+}}$. While the first objective term seeks to maximize the expected rewards, the second objective term incentivizes smoothness of the robot's trajectories.

\Cref{fig:pipeline} illustrates the trajectory planning process on the right panel. The VLM provides three action proposals. WoMAP optimizes each of the three action proposals, estimates their rewards, and then executes the optimized action sequence with the highest reward.

%% file: sections/experiments.tex
\section{Experiments}
\label{sec:experiments}
We evaluate WoMAP on open-vocabulary active object localization tasks both in simulation and on a TidyBot \cite{wu2024tidybot++} to answer the following questions: 
(1)~How does WoMAP perform across environments and tasks of varying difficulty (defined by scene complexity and initial conditions)? 
(2)~Can WoMAP transfer effectively from simulation to the real world when trained only on photorealistic data?
(3)~Can WoMAP achieve zero-shot generalization to visual (unseen lighting, backgrounds) and semantic (unseen instructions, target objects) conditions? 
Further, in Appendix~\ref{appendix:ablation-encoder}, we ablate training with image reconstruction objectives, freezing vs. finetuning pretrained image encoders, and training image encoders from scratch.

\begin{figure*}
    \centering
    \includegraphics[width=\linewidth]{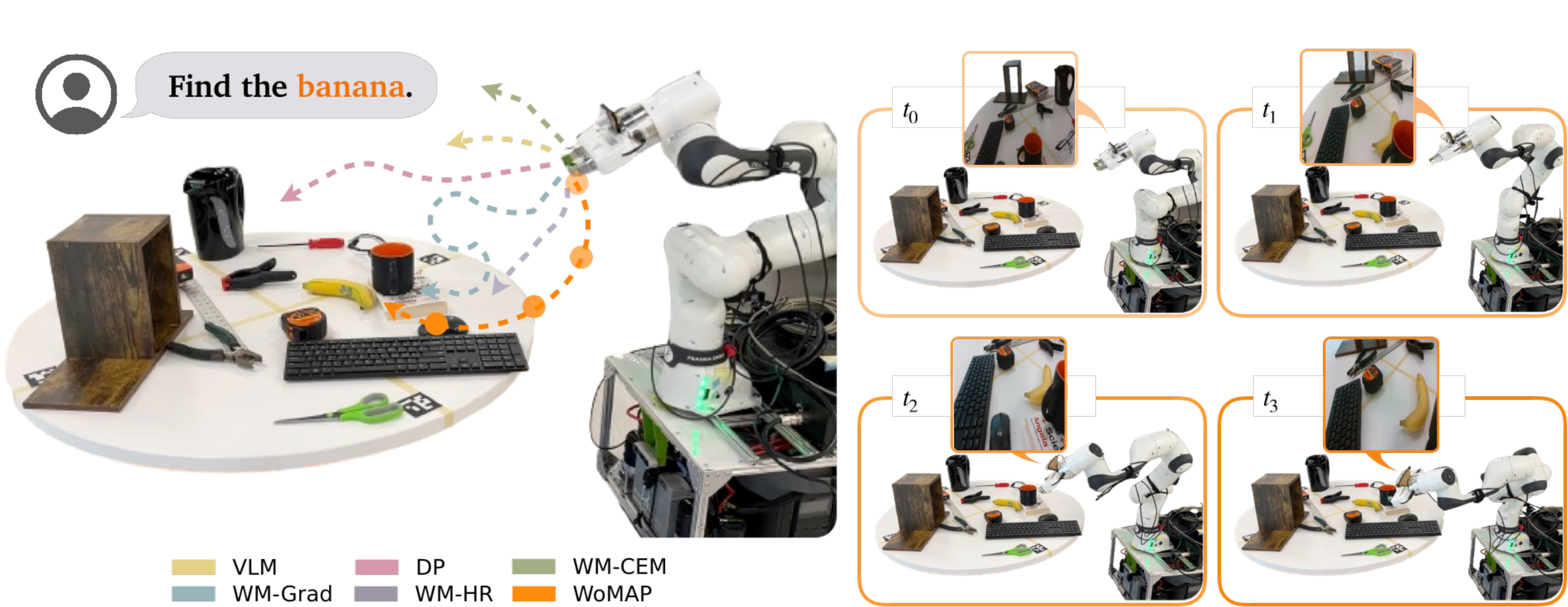}
    \caption{\textbf{Visualization of the TidyBot's trajectories for all planners.} When asked to find an object, e.g., a banana occluded by a mug, WoMAP finds the target object (banana) more efficiently than the other planners. As illustrated, the WM-Grad computes inefficient, circuitous paths, while the DP does not look behind occlusions. See Section~\ref{ssec:exp_eval_task_difficulty} and the paper's video for more details. 
    Further, we show images from the scene and wrist cameras at different timesteps when planning with WoMAP (right). 
}
    \label{fig:traj}
\end{figure*}

\subsection{Environments and Tasks}
\label{subsec:environments-tasks}
We consider four PyBullet (PB) simulation environments~\cite{coumans2020} and three real-world environments both within Gaussian Splats (GS) and on a TidyBot, focusing on practical scene configurations, e.g., with \emph{kitchen}, \emph{office}, and \emph{random} environment themes. 
Within each environment, we randomly vary object configurations and target identities to create challenging localization tasks with occlusions and distractors.
For a comprehensive evaluation of all methods, we vary the task difficulty along two axes: (i) \emph{scene difficulty}, determined by the number, diversity, and layout of objects and (ii) \emph{initial-pose difficulty}, representing the task difficulty due to occlusions, viewability, and distance to the target object which depends on the initial pose of the robot.
In Appendix~\ref{ssec:appendix_task_design}, we provide a detailed discussion of the evaluation setup, as well as illustrations of the tasks and environments.

\subsection{Baselines, Ablations, and Evaluation Metrics}
For baselines, we first benchmark WoMAP against a VLM-based planner using GPT-4o \cite{hurst2024gpt}, similar to \cite{goetting2024end, ren2024explore}, prompted with the same template as WoMAP (Appendix~\ref{ssec:appendix_womap_planning}).
We also evaluate a multi-task diffusion policy (DP), trained on expert demonstrations across multiple target objects for the same set of training environments.
Additionally, we compare against world model-only planners:
(i)~\emph{WM-CEM} \cite{zhou2024dino}, using the gradient-free, cross-entropy method for action proposals;
(ii)~\emph{WM-Grad}, using gradient-based action optimization;
and
(iii)~\emph{WM-HR}, a heuristic-based, non-VLM planner using a fixed set of atomic actions (the same set used to prompt the VLM), refined via gradient descent without VLM input.
We use success rate and efficiency (given by the success rate weighted by the path length \cite{anderson2018evaluation}) as evaluation metrics, where success is defined by a threshold on the detection confidence of the target object and the proportion of the associated bounding-box in the robot's camera observation.
See Appendix~\ref{ssec:appendix_baselines_ablations_metrics} for additional implementation details.

\subsection{Evaluation across Varying Task Difficulty}
\label{ssec:exp_eval_task_difficulty}
\begin{figure*}[t]
    \centering
    \includegraphics[width=1.0\linewidth]{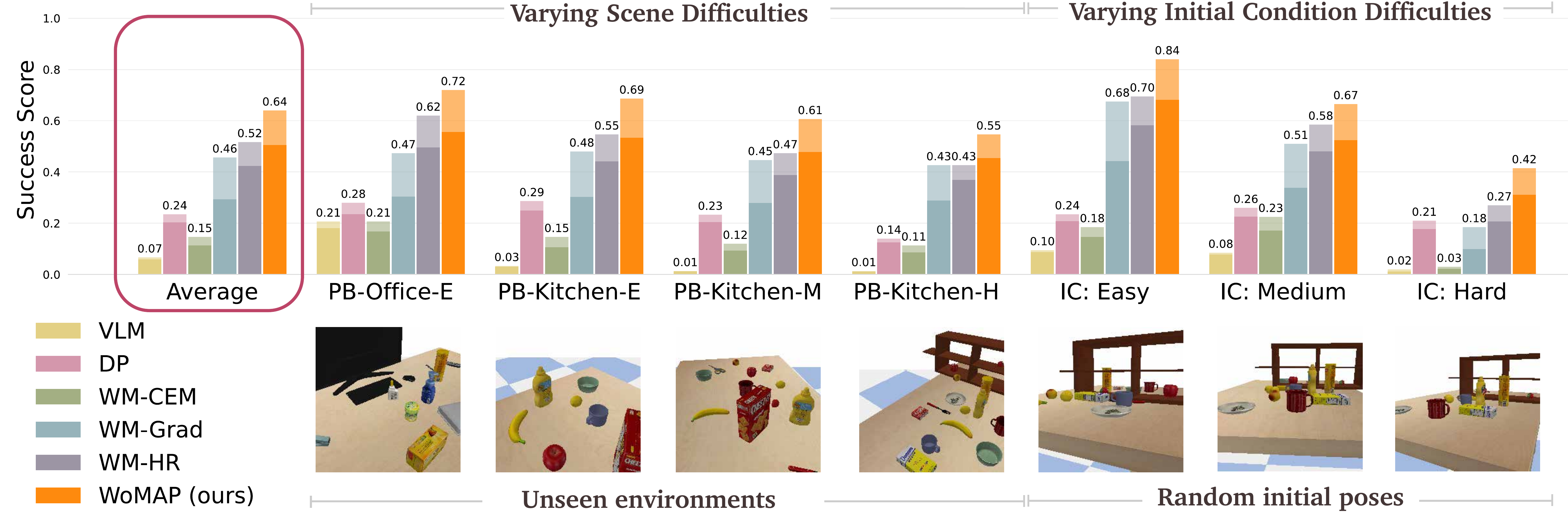}
    \caption{\textbf{PyBullet evaluation tasks and results.} 
    Success rates (translucent bars) and efficiency scores (solid bars) in active object localization across PyBullet scenes (presented in the order of increasing difficulty) and initial-pose conditions: easy (E), medium (M), and hard (H). WoMAP outperforms all baseline methods in all scenes and initial-pose conditions.}
    \label{fig:baselines_pybullet_plot_dp_binary_success}
\end{figure*}

\begin{figure*}[th]
    \vspace{-10pt}
    \centering
    \includegraphics[width=1.0\linewidth]{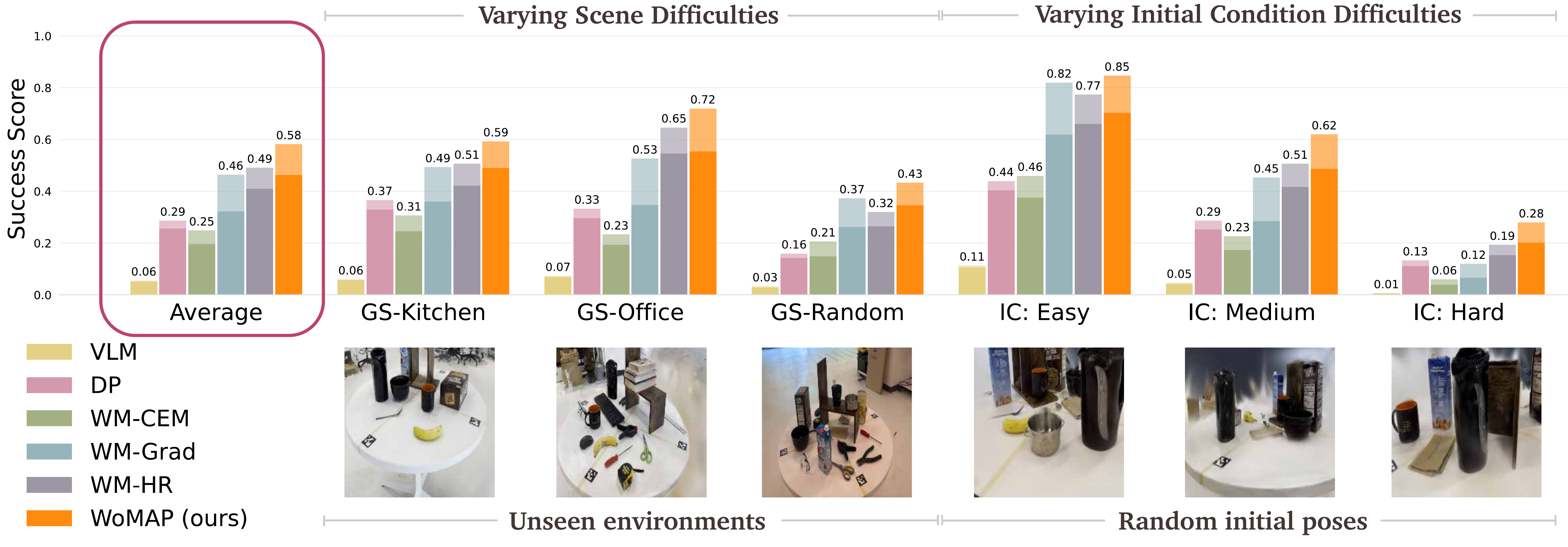}
    \caption{\textbf{Gaussian Splat evaluation tasks and results.} 
    Success rates (translucent bars) and efficiency scores (solid bars) in active object localization across Gaussian Splat scenes and initial-pose conditions: easy (E), medium (M), and hard (H). As in the PyBullet scenes, WoMAP outperforms all baseline methods via effective action grounding and optimization.}
    \label{fig:baselines_gsplat_plot_dp_binary_success}
\end{figure*}

We evaluate each method on novel (unseen) scenes in PyBullet and Gaussian Splat with varying scene and initial-pose difficulty, across a total of 150 tasks per environment.
\Cref{fig:baselines_pybullet_plot_dp_binary_success,fig:baselines_gsplat_plot_dp_binary_success} illustrate that on average WoMAP outperforms the VLM and diffusion policy (DP) baselines by over $9\times$ and $2\times$, respectively. 
While the VLM planner fails to account for physical grounding and the DP policy struggles to generalize beyond the training distribution, WoMAP generates grounded actions across diverse scenes. 
Moreover, we observe a progressive increase in the performance of the world-model-based planners with more informed search methods,
in the order of (i) sampling-based WM-CEM, which is sample-inefficient even with $4\times$ as many action proposals, (ii) gradient-based WM-Grad, which relies on myopic local gradients and generates inefficient, circuitous actions, evidenced by its much lower efficiency compared to success scores, (iii) heuristics-based WM-HR, which does not leverage intelligent guidance from the VLM, limiting its performance in challenging problems, and ultimately (iv) WoMAP, which evaluates and optimizes the VLM action proposals with the world model. %
\Cref{fig:traj} visualizes example trajectories from all planners on a TidyBot tasked with finding a banana hidden behind a mug, illustrating these results. Notably, the VLM fails to approach the target closely, while DP does not produce useful exploration behavior.
Next, we discuss the performance of the planners with respect to the difficulty of the scene and initial pose and direct readers to Appendix~\ref{sec:appendix-exp} for ablations on the correlation between data quantity/scene diversity and performance.

\textbf{Varying Scene Difficulty.}
As expected, the performance of all methods decreases with increasing scene difficulty, with a relative drop in success rates over $50\%$ for VLM and DP baselines from the easiest scene to the hardest. In comparison, WoMAP's performance only drops by $23.6\%$ in PyBullet and $40.2\%$ in Gaussian Splat. Notably, all world-model-based planners exhibit lower performance drops, suggesting the importance of planning with physical priors provided by world models.

\textbf{Varying Initial Conditions.} 
WoMAP achieves the second-smallest relative performance drop in the PyBullet environments (after DP) and the smallest performance drop in the Gaussian Splat environments, even with its highest absolute scores. %
By leveraging the high-level reasoning capabilities of the VLM for action proposals, WoMAP effectively mitigates hallucinations in world models that arise in difficult-to-predict scenarios, e.g., occlusions.

\newpage
\subsection{Sim-to-Real Transfer with Gaussian Splats}
\label{sec:experiments_sim_to_real_transfer}

\ifbool{rss_format}
{
\setlength{\tabcolsep}{1.2pt}
\begin{table}
    \centering
    \caption{Success rates ($\%$) for zero-shot sim-to-real transfer for VLM and WoMAP. }
    \label{tab:sim-to-real}
    \begin{tabular}{lccc}
    \toprule
    \textbf{Model} & \textbf{GS-Kitchen}  & \textbf{GS-Office}  &\textbf{GS-Random} \\
     \midrule
     VLM (sim) & $6$ & $13$ & $3$ \\
     VLM (real) & $0$ & $5$ & $0$ \\
     \midrule
     WoMAP (sim) & $71$ & $65$ & $32$ \\
     WoMAP (real) & $55$ & $65$ & $63$ \\
   \bottomrule 
    \end{tabular}
\end{table}
}
{
\setlength{\tabcolsep}{1.2pt}
\begin{wraptable}[7]{r}{6cm}
    \vspace{-20pt}
    \centering
    \scriptsize
    \centering
    \centering
    \caption{Success rates ($\%$) for zero-shot sim-to-real transfer for VLM and WoMAP. }
    \label{tab:sim-to-real}
    \begin{tabular}{lccc}
    \toprule
    \textbf{Model} & \textbf{GS-Kitchen}  & \textbf{GS-Office}  &\textbf{GS-Random} \\
     \midrule
     VLM (sim) & $6$ & $13$ & $3$ \\
     VLM (real) & $0$ & $5$ & $0$ \\
     \midrule
     WoMAP (sim) & $71$ & $65$ & $32$ \\
     WoMAP (real) & $55$ & $65$ & $63$ \\
   \bottomrule 
    \end{tabular}
\end{wraptable}
}

We evaluate WoMAP's sim-to-real transfer ability on 20 hardware trials for each of the 3 corresponding real-world tasks using the TidyBot.
For each trial, we randomize both the scene configuration and target object to include a diverse range of initial conditions and task difficulties.
As shown in Table~\ref{tab:sim-to-real}, despite being trained entirely in the Gaussian Splat simulation, WoMAP transfers effectively to the real world, achieving the same success rate in \emph{GS-Office} and a higher success rate in \emph{GS-Random} compared to the sim success rates, with only a moderate performance drop of $23\%$ in \emph{GS-Kitchen}. 
In contrast, the VLM baseline typically predicts unrealistic actions that violate joint limits, resulting in a substantial performance drop of $62\%$ or more.
This finding highlights WoMAP’s strong generalization capabilities in producing reliable reward predictions under domain shifts, enabling effective sim-to-real transfer.

\begin{figure*}[t]
    \centering
    \includegraphics[width=\linewidth]{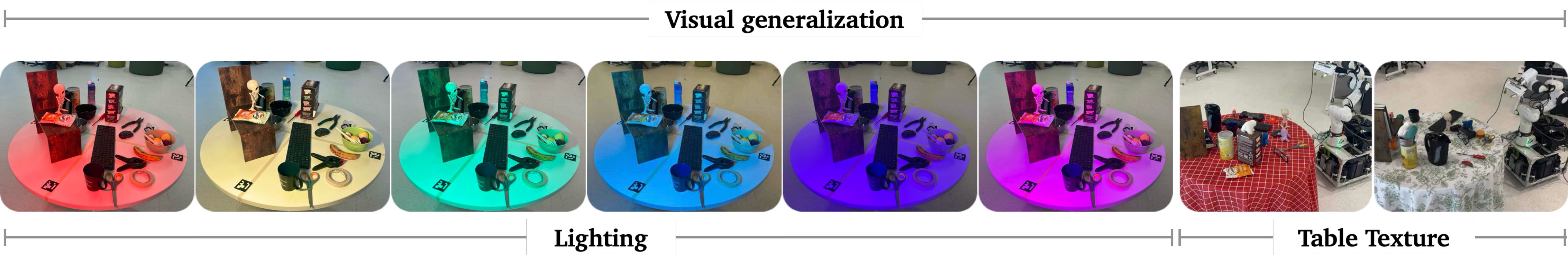}
    \caption{\textbf{Visual generalization setup:} lighting and background conditions.}
    \label{fig:exp_visual_generalization}
\end{figure*}

\subsection{Generalization to Novel Task Conditions}
\label{ssec:exp_generalization_to_novel_task_conditions}

\ifbool{rss_format}
{
\setlength{\tabcolsep}{1.2pt}
\begin{table}
    \centering
    \caption{Visual generalization results for various background and lighting conditions. }
    \label{tab:results_visual_generalization}
    \begin{tabular}{lcc}
    \toprule
    \textbf{Axis} & \textbf{Success Rate $\%$}  & \textbf{Efficiency $\%$} \\
     \midrule
     Nominal & $63$ & $60$ \\
     Lighting  & $50$ & $47$ \\
     Backgrounds & $30$ & $28$ \\
   \midrule 
    \end{tabular}
\end{table}
}
{
\setlength{\tabcolsep}{1.2pt}
\begin{wraptable}[6]{R}{6cm}
    \vspace{-15pt}
    \centering
    \scriptsize
    \centering
    \centering
    \caption{Visual generalization results for various background and lighting conditions. }
    \label{tab:results_visual_generalization}
    \begin{tabular}{lcc}
    \toprule
    \textbf{Axis} & \textbf{Success Rate $\%$}  & \textbf{Efficiency $\%$} \\
     \midrule
     Nominal & $63$ & $60$ \\
     Lighting  & $50$ & $47$ \\
     Backgrounds & $30$ & $28$ \\
   \midrule 
    \end{tabular}
\end{wraptable}
}

\begin{figure*}[t]
    \centering
    \includegraphics[width=\linewidth]{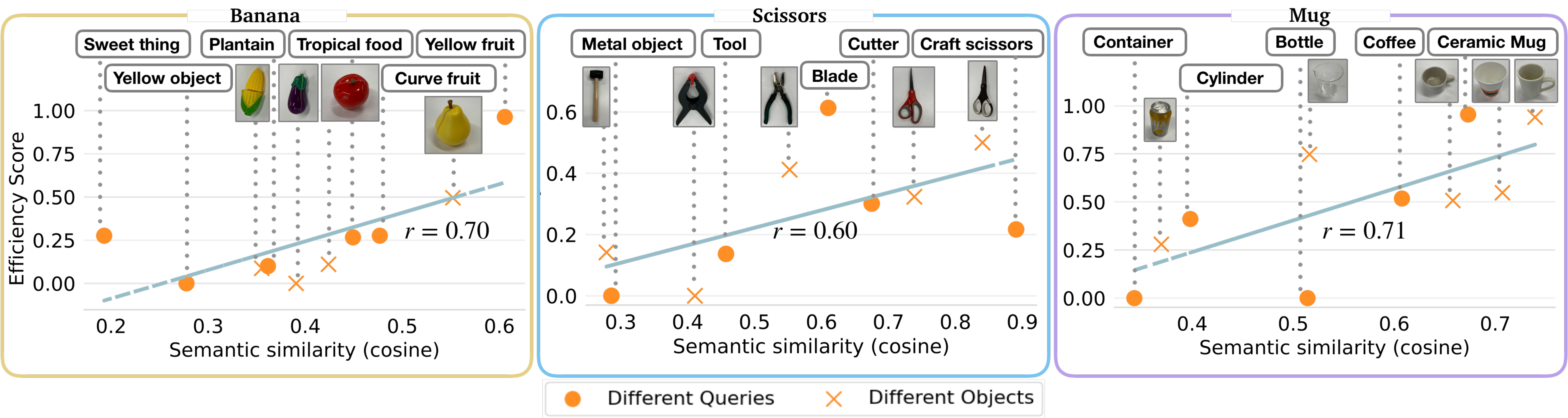}
    \caption{\textbf{Generalization plots for unseen queries and objects in the same category:} (left) \emph{banana}, (center) \emph{scissors}, (right) \emph{mug}. We see a positive correlation in semantic similarity (cosine distance) of the objects/queries with the most similar object present in our training objects, and the efficiency score suggesting the model's performance.
    }
    \label{fig:generalization-bsm}
\end{figure*}

We examine the visual and semantic generalization capabilities of WoMAP trained only on nominal conditions in \emph{GS-Random} on $10$ scenes with $30$ trajectories each. We evaluate WoMAP in out-of-distribution lighting and background conditions, illustrated in \Cref{fig:exp_visual_generalization}. %
In \Cref{tab:results_visual_generalization}, we show that WoMAP achieves strong zero-shot generalization with a success rate and efficiency score of $50\%$ and $47\%$ compared to $63\%$ and $60\%$ in nominal conditions, respectively (a decrease of less than $22\%$), even under extreme lighting conditions, with a further performance drop with out-of-distribution backgrounds. In general, these findings show that WoMAP learns robust latent-space features for generalization to out-of-distribution test-time conditions.

In addition, we evaluate semantic generalization of WoMAP to unseen target objects and task instructions across two axes: (i)~to unseen language instructions to find target objects that were seen during training and (ii)~to unseen target objects with unseen language instructions. As illustrated in \Cref{fig:generalization-bsm}, we consider three representative object categories: ``banana," ``scissors," and ``mug," where WoMAP is only trained on a single instance for each.
We find that WoMAP achieves strong semantic generalization, with an expected decrease in performance with decreasing \emph{semantic similarity} as measured by the the cosine similarity metric.
For example, we ask WoMAP to find  the following unseen objects: ``pear," ``pliers," and ``beaker." Even though WoMAP has not seen these objects during training, WoMAP is able to find each of these objects at test time. 
We discuss these results further in Appendix~\ref{sec:appendix-exp}.

%% file: sections/conclusion.tex
\section{Conclusion}
\label{sec:conclusion}
We present \algname, a recipe for open-vocabulary active object localization that uses a scalable data generation pipeline to train a latent world model without expert demonstrations or online interaction data. 
\algname distills dense reward signals into the world model with a reconstruction-free training architecture for strong generalization from a few training samples. At inference time, WoMAP utilizes the world model for dynamics and rewards prediction to ground high-level action proposals from VLMs, demonstrating more efficient object localization and strong generalization to novel tasks.

%% file: sections/limitations_future_work.tex
\section{Limitations and Future Work}
\label{sec:limitations_future_work}

\noindent\textbf{Interactive Active Object Localization.}
Although we limit our problem to non-interactive object localization problems in this work, interaction between the robot and its environment is crucial to efficient exploration in many problem setting. Consequently, active object localization with interactive feedback from the environment is a promising direction for future research to enable more expressive and manipulation-intensive tasks.

\smallskip
\noindent\textbf{Uncertainty Quantification in Rewards Distillation.}
WoMAP distills the confidence of pretrained object detectors into a world model as a rewards signal. However, learned object detectors sometimes produce uncalibrated confidence estimates, which could corrupt the training data, negatively impacting its effectiveness in grounding action proposals. 
Future work will explore calibration methods for pretrained object detectors to ensure data fidelity during training. 

\smallskip
\noindent\textbf{Hallucination Detection and Uncertainty Quantification in World Models.}
WoMAP's action optimization fails when the world model hallucinates the dynamics/rewards, usually in areas where the world model is not confident. Uncertainty quantification of world models remains critical to identifying when to trust predictions from world models, which has been relatively unexplored. In future work, we will derive calibrated uncertainty quantification methods to enable uncertainty-aware planning to ensure effective action grounding and optimization. In addition, we will explore incorporating calibrated uncertainty estimates from the VLM on the action proposals into WoMAP to enable risk-sensitive planning.

%% file: sections/final_appendix.tex
\input{sections/appendix-impl}

\input{sections/appendix-exp}

%% file: sections/appendix-impl.tex
\section{Implementation Details}
\label{sec:appendix-impl}

\subsection{Details on Data Generation}
\subsubsection{Preliminaries: Gaussian Splatting for Photorealistic Novel View Rendering}

We provide a brief introduction to Gaussian Splatting and its applications to robotics.
Gaussian Splatting \cite{kerbl20233d} is a volumetric scene representation that uses explicit ellipsoidal primitives to represent non-empty space in any given environment. Trained entirely from RGB poses and associated camera poses, Gaussian Spatting enables real-time photorealistic novel-view synthesis without any structural priors, unlike many existing scene reconstruction methods.
Given its high-fidelity reconstruction, novel-view synthesis, and amenability to open-vocabulary semantics, Gaussian Splatting has been widely applied in robotics, e.g., robot manipulation \cite{shorinwa2024splat, qureshi2024splatsim}. Training a latent world model requires abundant data coverage of the environment, which is challenging to collect in the real world. In this work, we leverage Gaussian Splatting for scalable data generation from only a few real-world videos. Specifically, we employ semantic Gaussian Splatting \cite{zhou2024feature, shorinwa2024fast} for automatic labeling of target objects, distilling language semantics from CLIP \cite{radford2021learning} into the Gaussian Splat. In the following subsections, we briefly describe the procedure used for generating, aligning, and rendering views for multiple scenes. 

\textbf{Collecting Videos and Training the Gaussian Splat.} Trained Gaussian Splats do not share a common reference frame, in general. Hence, we align the individual Gaussian Splats using four Aruco markers in fixed positions. However, we note that other approaches such as semantics-based alignment can also be used. In each scene, we record a one-minute video as input to the Gaussian Splat.
We compute the camera poses for each video using structure-from-motion and subsequently train the semantic $3$D Gaussian Splat \cite{shorinwa2025siren} for 30,000 iterations on an Nvidia L40 GPU using Nerfstudio \cite{tancik2023nerfstudio}.

\begin{figure*}[h]
    \centering
    \includegraphics[width=\linewidth]{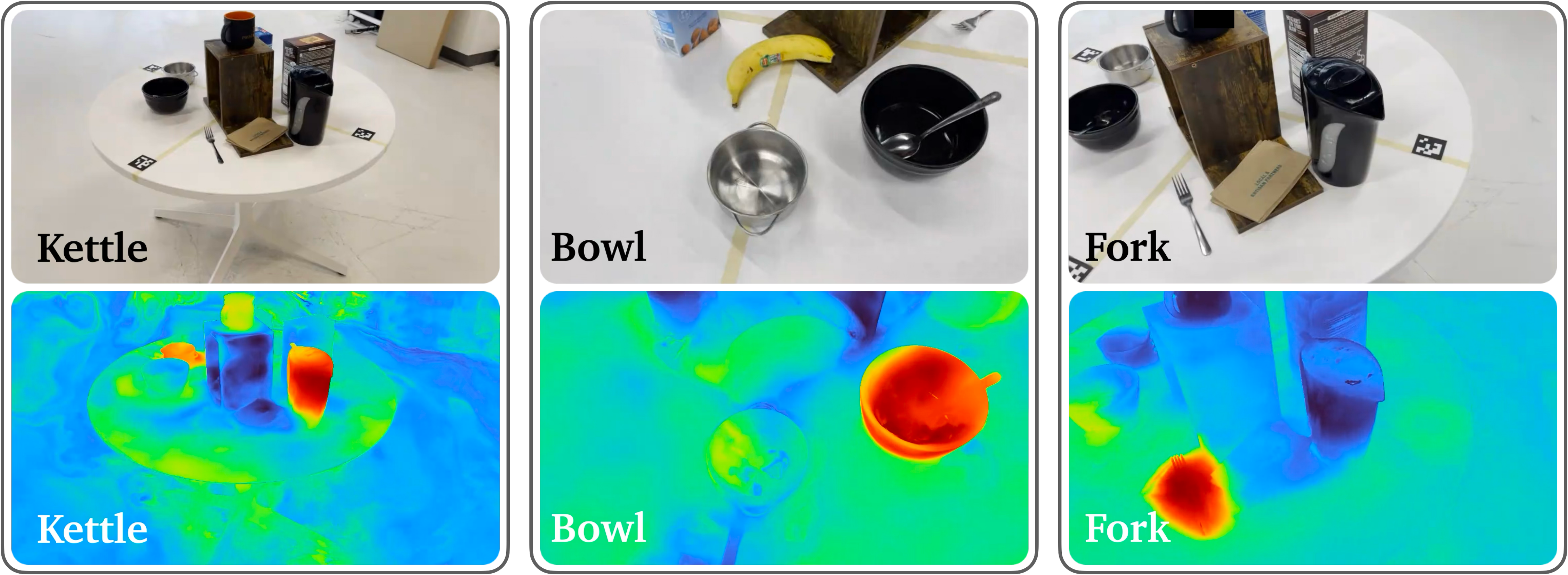}
    \caption{Querying the semantic field of the Gaussian Splat.}
    \label{fig:gsplat-vis}
\end{figure*}

\textbf{Scene Alignment and Annotation. } The reconstructed scene representation could have an arbitrary reference frame. However, a common reference frame is necessary for data consistency when generating data from multiple scenes.
Consequently, we first perform alignment of the scene coordinates by matching Aruco tags detected in the reconstruction with its measured ground-truth position and orientation. 
Finally, given coordinates of the detected points in the world frame, we solve the Perspective-n-Point (PnP) problem and the point-registration problem using RANSAC to compute the camera-to-world and the Gaussian Splat-to-world transforms, respectively.  
We query the semantic field of the Gaussian Splat to annotate the position and dimension of each target object in the scene, visualized in Figure~\ref{fig:gsplat-vis}, with different target objects.

\subsubsection{Ablation: Generating Training Data without Novel Views}
We examine the need for data scalability beyond real-world data, ablating WoMAP limited to real-world video frames compared to the data generated from the Gaussian Splats. First, we extract a set of image frames uniformly across each real-world video and use these images in training the Gaussian Splat. Next, we train two world models with: (i)~only the video frames of each scene (\emph{Video-Only}) and (ii)~on the rendered images from the Gaussian Splat (\emph{GSplat-Data}). 
The training data of the Video-Only and GSplat-Data world models consist of about $2100$ and $9000$ images, respectively. We evaluate the trained model using the gradient-based planner WM-Grad and summarize our results in Table~\ref{tab:results-sim2real}.
Across all scenes and initial conditions, the GSplat-Data model outperforms the Video-Only model by significant margins, ranging between $50\%$ and $200\%$. The relatively poor performance of the Video-Only model can be explained by the lack of sufficient data coverage in the real-world videos, underscoring the importance of our scalable data pipeline. Our data generation pipeline provides not only additional training images, but also \emph{diverse} viewpoints, which makes the GSplat-Data model more robust to initial conditions compared to the Video-Only model (sweeping from \emph{easy} to \emph{hard}).

\setlength{\tabcolsep}{4pt}
\begin{table*}[h]
    \centering
    \caption{
    Success scores of the world model trained using only real-world video frames (Video-Only model) compared to the (GSplat-Data model). With more diverse data, the GSplat-Data model outperforms Video-Only model across scenes and initial conditions.}
    \label{tab:results-sim2real}
    \begin{adjustbox}{width=\linewidth}{
    \begin{tabular}{lccc|ccc|ccc}
    \toprule
    \textbf{Training Data} 
    & \multicolumn{3}{c|}{\textbf{GS-Kitchen}} 
    & \multicolumn{3}{c|}{\textbf{GS-Office}} 
    & \multicolumn{3}{c}{\textbf{GS-Random}} \\
    & init-easy & init-medium & init-hard
    & init-easy & init-medium & init-hard
    & init-easy & init-medium & init-hard \\
    \midrule
    Video-Only & $0.06$ & $0.04$ & $0.02$ & $0.24$ & $0.06$ & $0.02$ & $0.04$ & $0$ & $0$ \\
    GSplat-Data & $0.84$ & $0.50$ & $0.14$ & $0.86$ & $0.50$ & $0.22$ & $0.76$ & $0.36$ & $0$ \\
    \bottomrule
    \end{tabular}}
    \end{adjustbox}
    \vspace{3pt}
\end{table*}

\subsubsection{Details on Trajectory Data Generation}
\label{appendix-impl-data-gen}

\begin{figure*}[th]
    \centering
    \includegraphics[width=0.8\linewidth]{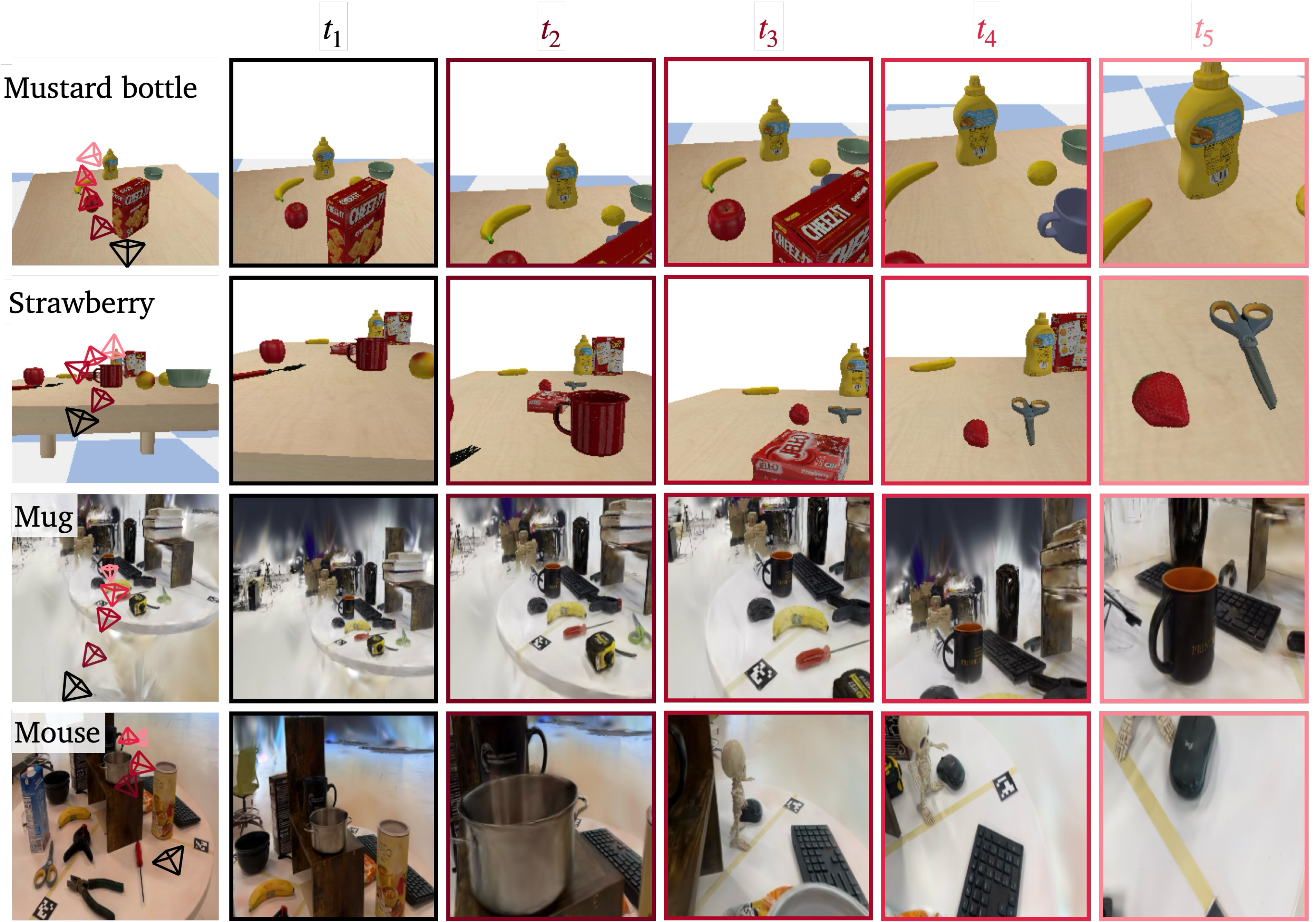}
    \caption{Visualization of training trajectories generated in PyBullet and Gaussian Splat.
    }
    \label{fig:training-data}
\end{figure*}

To get sufficient coverage of diverse viewpoints in the the PyBullet environment and Gaussian Splat scenes, we generate synthetic collision-free trajectories that start from randomized initial positions and navigate towards various target locations.
Specifically, we leverage the RRT* planner \cite{lavalle2001rapidly} to compute feasible, diverse paths between the start and goal. We concatenate these camera poses and add random linear and angular perturbations to further increase data diversity to cover a more realistic range of viewpoints encountered at deployment time. Figure~\ref{fig:training-data} provides a visualization of trajectories generated in our training dataset. We collect observations at relatively low frequencies where the delta distance between consecutive observations is around 1-5cm. The largest model that we train contains about 10,000 observation-camera pose pairs for 500 trajectories (or 20 observations per trajectory), which is considerably smaller in scale than the training data used in many other imitation learning or reinforcement learning-based visual navigation policies \cite{shah2023navigationlargelanguagemodels, sridhar2024nomad}. Moreover, we show in Appendix~\ref{appendix:training-data-ablation} that WoMAP's performance remains competitive in much smaller training configurations.

\subsection{World Model Implementation}
\label{ssec:appendix_world_model_implementation}
Here, we summarize the implementation details of the world model---composed of the observation encoder, dynamics predictor, and rewards predictor---and provide the hyperparameters used in training the world model. For interpretability, we train a decoder to map latent states to the image space without backpropagating the gradients through the other components of the world model. In addition, to better visualize the objects the world model is focusing on, we train the rewards models to predict bounding-boxes along with the scalar rewards, which we overlay in the decoded RGB images.

\noindent\textbf{Observation Encoder.}
We encode raw image observations into the latent space using the pre-trained DINOv2 model \emph{dinov2\_vits14} \cite{oquab2023dinov2}. We use the norm of the patch tokens of image $o_{t}$ as its feature embedding ${h_{\theta}(o_{t}) \in \mbb{R}^{384}}$. In addition, we freeze the weights of the DINOv2 model during training and do not apply any image data augmentation, e.g., color jittering and random cropping, since the pre-trained DINOv2 model already utilizes data augmentation for training. Moreover, random perturbation of the image, such as random rotation or cropping, may compromise the fidelity of the ground-truth image-action pairs. In Appendix~\ref{appendix:ablation-encoder}, we ablate the observation encoder.

\smallskip
\noindent\textbf{Dynamics Predictor.}
To condition the ViT-based dynamics predictor on both the latent state $z_{t}$ and action $a_{t}$, we map $z_{t}$ and $a_{t}$ to a $384$-dimensional embedding space using an affine transformation and concatenate the resulting embeddings. 
In preliminary experiments, we found that a longer historical context for dynamics prediction did not provide any significant improvement in prediction accuracy. As a result, we provide only the last latent state to the dynamics predictor.
To improve the multi-step prediction accuracy, we supervise the dynamics predictor over a sequence of observation-action pairs, recursively passing in the previous prediction into the model.
Consequently, we do not optimize the dynamics predictor using teacher forcing \cite{williams1989learning}. Although teacher forcing facilitates faster training through parallelism, teacher forcing contributes to significant accumulation of dynamics errors over multi-step predictions.

\smallskip
\noindent\textbf{Rewards Predictor.}
Like the dynamics predictor, we condition the rewards predictor on $z_{t}$ and the language embedding $e_{g}$, each mapped to $\mbb{R}^{384}$. We concatenate the embeddings and apply full cross-attention to predict the scalar reward for that latent state and target object. We train the rewards predictor using the binary cross-entropy loss function given by: ${\ell_{t} = - [ r_{\text{gt}, t} \log(r_{t})+ (1 - r_{\text{gt}, t}) \log(1 - r_{t}))  ]}$, for datapoint ${(r_{\text{gt}, t}, r_{t})}$ with ground-truth reward $r_{\text{gt}, t}$ and predicted reward $r_{t}$.

\smallskip
\noindent\textbf{Training Setup and Hyperparameters.}
We train the world model on a single Nvidia L40 GPU with 48GB of GPU VRAM using a batch size of $25$ for between $8$ to $10$ hours, depending on the task environment. In \Cref{tab:womap_world_model_hyperparams}, we present the hyperparameters used in training the world model. We warmup training with a learning rate (LR) of $1e^{-3}$ before training for the full number of epochs ($100$) with an LR of $5e^{-4}$ for stable training. We observed that the training loss diverged for learning rates greater than $1e^{-3}$.
Further, in \Cref{tab:womap_world_model_num_params}, we report the number of trainable and non-trainable parameters in each component of the world model. The dynamics $q_{\psi}$ and rewards $v_{\phi}$ predictors have about $7.6$ and $4.1$ million trainable parameters, respectively. Meanwhile, we do not fine-tune the observation encoder $h_{\theta}$.

\begin{table}[h]
    \setlength{\tabcolsep}{8pt}
    \parbox{0.48\textwidth}{
        \caption{WoMAP's Hyperparameters.}
        \label{tab:womap_world_model_hyperparams}
        \centering
        \begin{tabular}{l c}
            \toprule
            Name & Value \\
            \midrule
             Image Size & 224 \\
             Lang. Embed Dim & 384 \\
             Pred. Embed Dim & 384 \\
             Start LR & 1e-3 \\
             LR & 5e-4 \\
             Warmup Epochs & 2 \\
             Weight Decay & 4e-2 \\
             Final Weight Decay & 0.4 \\
             Batch Size & 25 \\
             Total Epoch & 100 \\
             Planning Horizon & 4 \\
            \bottomrule
        \end{tabular}
    }
    \hfill
    \setlength{\tabcolsep}{5pt}
    \parbox{0.48\textwidth}{
        \caption{WoMAP's Number of Parameters (in millions).}
        \label{tab:womap_world_model_num_params}
        \centering
        \begin{tabular}{l c c}
            \toprule
            Name & \# trainable & \# non-trainable \\
            \midrule
             $h_{\theta}$ & 0 & 22 \\
             $q_{\psi}$ & 7.6 & 0.1 \\
             $v_{\phi}$ & 4.1 & 0.1 \\
            \bottomrule
        \end{tabular}
    }
\end{table}

\subsubsection{Ablations}
\label{appendix:ablation-encoder}
\begin{figure*}[th]
    \vspace{-10pt}
    \centering
    \includegraphics[width=1.0\linewidth]{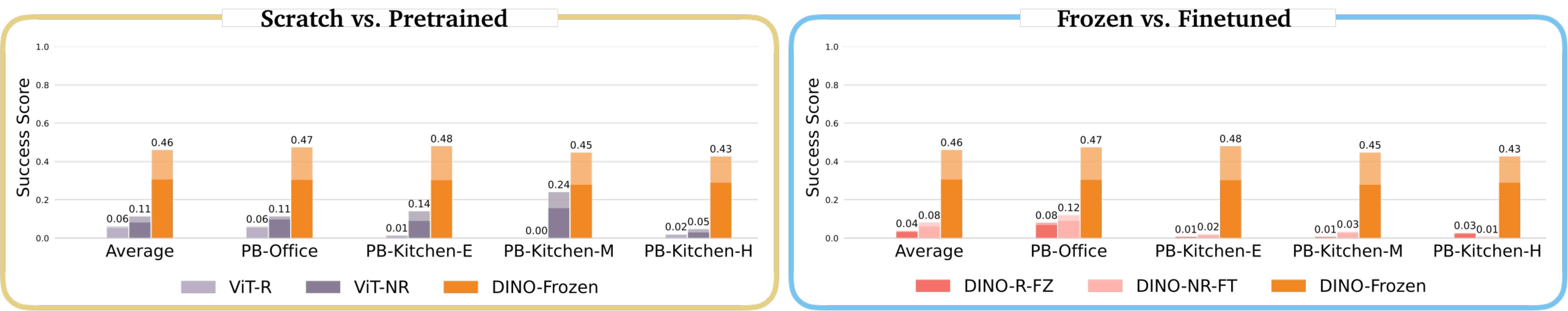}
    \caption{\textbf{World Model Architecture Ablations in the PyBullet Scenes.} 
    We explore training the observation encoder from scratch, finetuning, and training the observation encoder with image reconstruction, where \emph{ViT-R} denotes a ViT trained from scratch \textbf{with} image reconstruction, \emph{ViT-NR} denote a ViT trained from scratch \textbf{without} image reconstruction, respectively, \mbox{\emph{DINO-Frozen}} denotes a \textbf{frozen} DINOv2 encoder, \mbox{\emph{DINO-R-FT}} denotes a finetuned DINOv2 model trained \textbf{with} image reconstruction, and \emph{DINO-NR-FT} denotes a \textbf{finetuned} DINOv2 encoder trained \textbf{without} image reconstruction. WoMAP uses DINO-Frozen.
    }
    \label{fig:ablations_pybullet_eval_task_difficulty}
\end{figure*}

\begin{figure*}[th]
    \vspace{-10pt}
    \centering
    \includegraphics[width=1.0\linewidth]{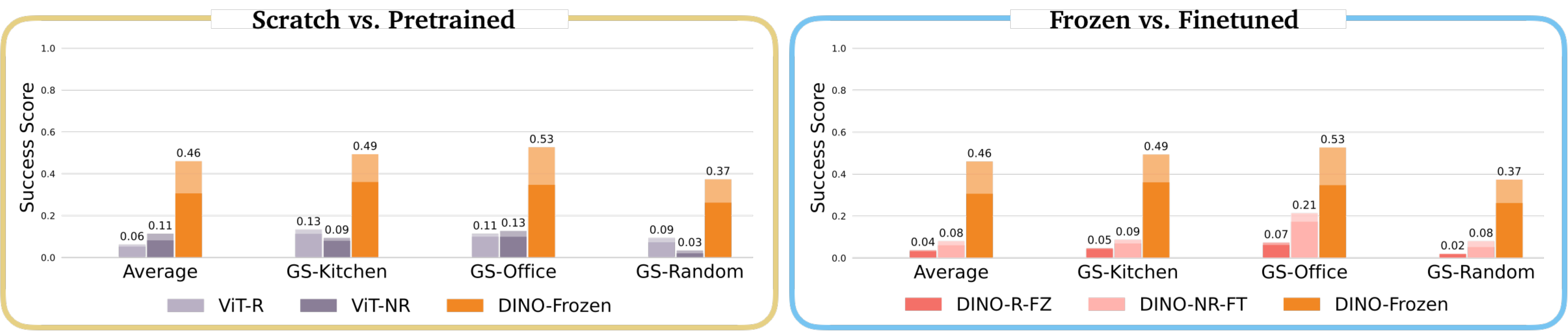}
    \caption{\textbf{World Model Architecture Ablations in the Gaussian Splatting Scenes.} 
    We ablate training the observation encoder from scratch, finetuning, and training the observation encoder with image reconstruction, where \emph{ViT-R} denotes a ViT trained from scratch \textbf{with} image reconstruction, \emph{ViT-NR} denote a ViT trained from scratch \textbf{without} image reconstruction, respectively, \emph{DINO-Frozen} denotes a \textbf{frozen} DINOv2 encoder, \emph{DINO-R-FT} denotes a finetuned DINOv2 model trained \textbf{with} image reconstruction, and \emph{DINO-NR-FT} denotes a \textbf{finetuned} DINOv2 encoder trained \textbf{without} image reconstruction. WoMAP uses DINO-Frozen.
    }
    \label{fig:ablations_gsplat_eval_task_difficulty}
\end{figure*}

\begin{figure*}[th]
    \vspace{-10pt}
    \centering
    \ifbool{rss_format}
    {
    \newcommand{\figwidth}{0.9}
    }
    {
    \newcommand{\figwidth}{1.0}
    }
    \includegraphics[width=\figwidth\linewidth]{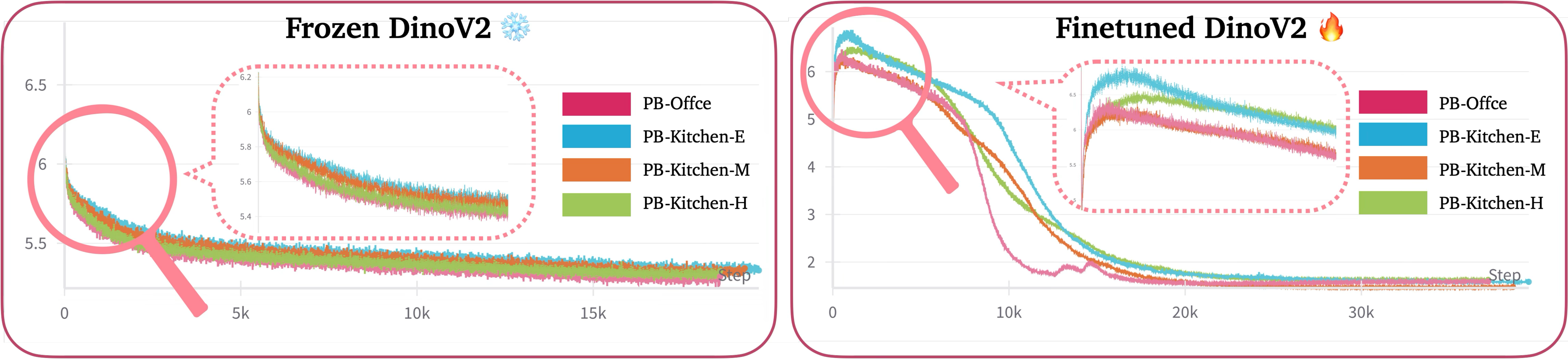}
    \caption{\textbf{Frozen vs. Finetuned DINOv2 Encoder.} 
    Finetuning the DINOv2 encoder generally leads to training instability, negatively impacting performance.
    }
    \label{fig:ablations_frozen_finetuned_dino}
\end{figure*}

We ablate different components of the world model, examining the effects of training an encoder from scratch, finetuning a pre-trained encoder, and using an image reconstruction loss. We report our findings in~\Cref{fig:ablations_pybullet_eval_task_difficulty},~\ref{fig:ablations_gsplat_eval_task_difficulty},~and~\ref{fig:ablations_frozen_finetuned_dino}, where \emph{ViT-R} and \emph{ViT-NR} denote ViT trained from scratch with image reconstruction and without image reconstruction, respectively, \emph{DINO-Frozen} denote a frozen DINOv2 model, and \emph{DINO-R-FT} and \emph{DINO-NR-FT} denote a finetuned DINOv2 model trained with and without image reconstruction, respectively. In all applicable plots, we represent the success rate of each method by the solid-color bars and the efficiency scores by the translucent bars. We discuss these results in the following subsections.

\noindent\textbf{Training the Observation Encoder from Scratch.}
We compare a ViT-based observation encoder trained from scratch (ViT-NR) to a frozen pre-trained DINOv2 model (DINO-NR-FZ), without an image reconstruction objective. From ~\Cref{fig:ablations_pybullet_eval_task_difficulty}~and~\ref{fig:ablations_gsplat_eval_task_difficulty}, DINO-NR-FZ achieves higher success rates and efficiency scores across the PyBullet and Gaussian Splat scenes. Although both models have access to the same data when training the world model, the results suggest that DINO-NR-FZ model benefits from large-scale pre-training, which provides a robust latent state for dynamics and rewards prediction even without any finetuning. 
Further, we observed that the ViT was more unstable to train, given the total number of trainable parameters.
In general, the ViT-NR and ViT-R models may require more training data to learn more useful visual features compared to the frozen DINOv2 models.

\noindent\textbf{Finetuning the Pre-trained Observation Encoder.}
We explore finetuning the DINOv2 encoder, comparing its performance to that of the frozen model. We find that finetuning the DINOv2 encoder leads to training instability that adversely impacts the performance of the world model. In fact, in many of our experiments, the training loss failed to decrease or raised Not-a-Number (NaN) errors. In ~\Cref{fig:ablations_frozen_finetuned_dino}, we show the training loss for the dynamics predictor across the four PyBullet environment, highlighting the increase in the training loss at the initial stages of the training procedure in the fine-tuned DINOv2 model. This training instability may be attributed to the more complicated loss landscape with many local minima during finetuning. In contrast, the training loss for the frozen DINOv2 models decreases relatively monotonically. These training dynamics are reflected in the success rates and efficiency scores achieved by both models. The finetuned models DINO-NR-FT and DINO-R-FT have notably lower scores on the performance metrics compared to frozen DINO encoder \emph{DINO-Frozen}.

\noindent\textbf{Training with a Reconstruction Objective.}
Here, we investigate training the world model with an image reconstruction objective. From ~\Cref{fig:ablations_pybullet_eval_task_difficulty,fig:ablations_gsplat_eval_task_difficulty}, 
we find that training with an image reconstruction objective generally leads to a degradation in the performance of the world model, e.g., in the ViT-R and DINO-R-FT models. Although image reconstruction objectives can provide dense rewards supervision, training instability often eliminates this potential advantage, underscoring the challenge with image reconstruction-based training. In some cases, the training instability can lead to significant drops in performance, e.g., in the ViT-R model when trained in the PB-Kitchen-M scene. 
In summary, the results from the ablations indicate that the frozen pretrained DINOv2 provides generalizable latent features that enable accurate dynamics and rewards prediction and effective action grounding.

\subsection{Details on Planning Integration}
\label{ssec:appendix_womap_planning}

\subsubsection{Planning with VLMs}

Using the world model as a local planner poses many challenges given the noisy reward landscape, particularly in the low training data regime. However, world models can serve as good evaluators/optimizers of imperfect action proposals generated from another policy. For example, generating good
6D action proposals with VLMs is difficult due to the limited spatial understanding ability of
VLMs, a challenge that can be addressed using a world model. We experimented with the following prompting strategies using GPT-4o \cite{hurst2024gpt}:

\textbf{Direct 6D action output.} We first provide rough dimensions of the scene and ask the VLM to directly output 6D actions (x, y, z, roll, pitch, yaw). However, due to inconsistencies in coordinate system conventions across its pretraining data (e.g., variations in the orientation of the positive x and y axes), the vision-language model often fails to consistently interpret spatial directions and rotations correctly.

\textbf{Direction output.} To minimize confusion on the coordinate axes definition, we experimented with expressing the translation and rotation actions with natural language descriptions, such as move forward/backward, tilt up/down, etc. However, we have found that these directions are not fully descriptive of exploration behavior. For example, if the VLM suggests looking behind an obstacle that is directly in front, its action outputs are often axis-aligned, which is not expressive enough to encode this ``look around" behavior.

\textbf{Action Primitives.} Our final version frames the prompt as a multiple-choice question, providing the VLM with a list of action primitives. Empirically, we find that using the coarse action primitives as shown in Fig. \ref{fig:vlm_prompt}, the VLM is capable of matching its high-level suggestions with the correct action outputs consistently. Despite covering only a limited set of actions with coarse magnitudes, we demonstrate that WoMAP can still successfully optimize these action proposals into grounded actions, as shown in \Cref{fig:pipeline}.

\begin{figure*}
\begin{center}
\begin{minipage}{0.98\textwidth}
\small
\begin{verbatim}
This is what you currently see. Please carefully analyze the current observation 
and think about what is the best action to take given what you see. 
If you think the current observation is a good enough view of {self.target} 
and you can't get any closer, please say `DONE'. 
Otherwise, please select the top {self.k} choices from the action options 
below that you think would help you achieve the goal given the current observation. 
The options are: 
    (A) Move directly forward for 15 cm -- this lets you approach the objects in view
    (B) Move directly to the left for 15 cm -- this expands your left view by a bit
    (C) Move directly to the right for 15 cm -- this expands your right views by a bit
    (D) Look to the left by 45 degrees -- this lets you look around
    (E) Look to the right for 45 degrees -- this lets you look around
    (F) Move forward-left for 15 cm -- this lets you approach objects on the 
    left side of your view
    (G) Move forward-right for 15 cm -- this lets you approach objects on the 
    right side of your view
    (H) Move forward-left for 15 cm and then look right by 45 degrees -- this lets 
    you look behind an object
    (I) Move forward-right for 15 cm and then look left by 45 degrees -- this lets 
    you look behind an object
All these actions should be executed relative to the current position.
Please think carefully step by step and reason about why your choice could help you
get the desired observation. 
Please also review your movement history to see where you've already explored. 
Output the results in a structured JSON format as follows: 
{ 
    "descriptions": <what you observe in the the current scene and 
    whether there are hints of the {self.target}.>
    "actions": [ 
        {"rank": 1, "choice": <choice1>, "confidence": <confidence_score1>, 
        "explanation": <explanation1>}, 
        ... 
        {"rank": {self.k}, "choice": <choice{self.k}>, "confidence": 
        <confidence_score{self.k}>, "explanation": <explanation{self.k}>}, 
    ] 
} 
<choice1>, ... <choice{self.k}> should be a single letter representing one of the 
9 choices above. 
Ensure the confidence scores are in descending order. 
Do not include any extra explanation outside of the JSON structure. 
\end{verbatim}
\captionsetup{type=figure}
\captionof{figure}{Prompt provided to the VLM}
\label{fig:vlm_prompt}
\end{minipage}
\end{center}
\end{figure*}

%% file: sections/appendix-exp.tex
\section{Experiment Details}
\label{sec:appendix-exp}

\subsection{Task Design Details}
\label{ssec:appendix_task_design}
We provide more details on the design choices for the tasks we examine WoMAP on.

\subsubsection{Designing task environments}
\begin{figure*}
    \centering
    \ifbool{rss_format}
    {
    \newcommand{\figwidth}{0.9}
    }
    {
    \newcommand{\figwidth}{1.0}
    }
    \includegraphics[width=\figwidth\linewidth]{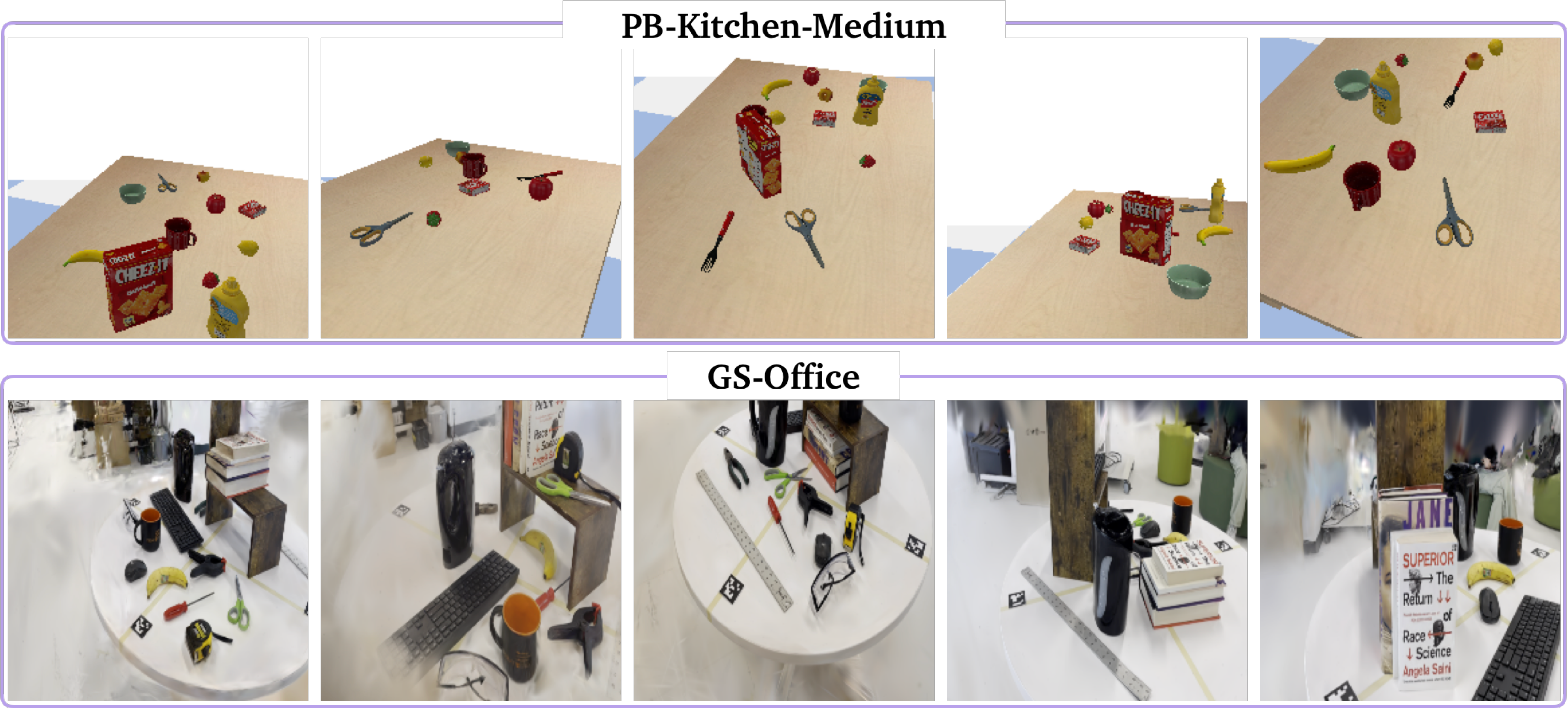}
    \caption{Scene variations for a single environment. We create multiple scenes within each environment by varying the configuration and degree of occlusions of the objects in the environment. We show a few scenes in the PB-Kitchen-Medium and GS-Office environments.}
    \label{fig:scene-variation}
\end{figure*}

As discussed in Section~\ref{subsec:environments-tasks}, we evaluate WoMAP on four simulation environments and three real environments. Figures~\ref{fig:baselines_pybullet_plot_dp_binary_success} and~\ref{fig:baselines_gsplat_plot_dp_binary_success} show example scenes for each environment, and Figure~\ref{fig:scene-variation} shows the different variations for a single environment across different scenes.

Our key motivation to selecting the suite of environments and objects lies on two axes: practicality and difficulty. We design each task based on a particular theme resembling a cluttered living area where active object localization is a a core challenge requiring spatial reasoning, viewpoint planning, and the ability to handle occlusions. 
For each environment, we pick representative objects along the axis of difficulty and try to cover a diverse set of objects. 
For \textit{easy} scenes, we pick distinct objects, which are typically small in size, whereas for difficult scenes, we place large shelves and visually similar objects (e.g., apple and peach), creating occlusions and distractions.
We provide a comprehensive list of environments and target objects in each environment in Table~\ref{tab:objects}. 

\begin{table*}[h]
\caption{Target objects used for each environments.}
\label{tab:objects}
\centering
\setlength{\tabcolsep}{4pt}
\begin{tabular}{>{\raggedright\arraybackslash}p{0.13\textwidth}
                >{\raggedright\arraybackslash}p{0.13\textwidth}
                >{\raggedright\arraybackslash}p{0.13\textwidth}
                >{\raggedright\arraybackslash}p{0.13\textwidth}
                >{\raggedright\arraybackslash}p{0.13\textwidth}
                >{\raggedright\arraybackslash}p{0.13\textwidth}
                >{\raggedright\arraybackslash}p{0.13\textwidth}}

\toprule
\textbf{PB-Kitchen-Easy} & \textbf{PB-Kitchen-Medium} & \textbf{PB-Kitchen-Hard} & \textbf{PB-Office} & \textbf{GS-Kitchen} & \textbf{GS-Office} & 
\textbf{GS-Random} \\
\midrule
 banana& banana& banana& gum& banana& banana&banana\\
 apple& apple& apple& lipton tea& mug& mug&mug\\
 green bowl& green bowl& green bowl& small marker& bowl& scissors&bowl\\
 lemon& lemon& lemon& glue& fork& books&fork\\
 mustard& mustard& mustard& book& pot& keyboard&keyboard\\
 cracker box& cracker box& sugar box& stapler& kettle& mouse&mouse\\
 blue cup& mug& mug& mug& & screwdriver&screwdriver\\
 & scissors& scissors& scissors& & eyeglass&eyeglasses\\
 & peach& peach& remote& & &pot\\
 & fork& fork& cleanser& & &scissors\\
 & strawberry& strawberry& potato chip& & &milk box\\
 & jello box& jello box& & & &skeleton\\
 & & plate& & & &\\
 & & potato chip& & & &\\
\bottomrule
\end{tabular}
\end{table*}

\subsubsection{Evaluating Task Difficulty}

\begin{figure*}
    \centering
    \ifbool{rss_format}
    {
    \newcommand{\figwidth}{0.9}
    }
    {
    \newcommand{\figwidth}{1.0}
    }
    \includegraphics[width=\figwidth\linewidth]{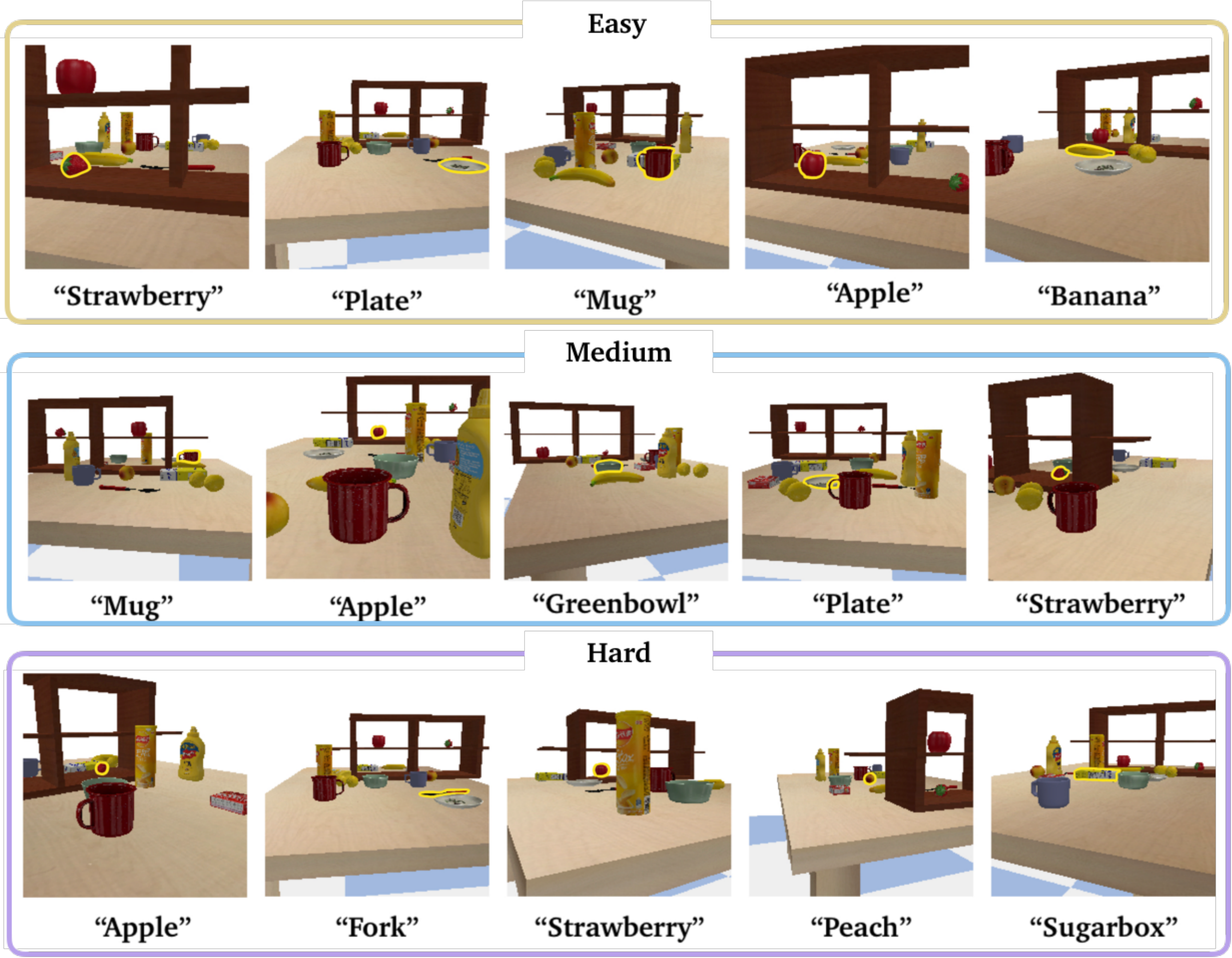}
    \caption{\textbf{PB-Kitchen-Hard initialization difficulty.} Initial observations and target query for different initial-pose difficulty levels in the PB-Kitchen-Hard scene.
    }
    \label{fig:pb-init-exp}
\end{figure*}

\begin{figure*}
    \centering
    \ifbool{rss_format}
    {
    \newcommand{\figwidth}{0.9}
    }
    {
    \newcommand{\figwidth}{1.0}
    }
    \includegraphics[width=\figwidth\linewidth]{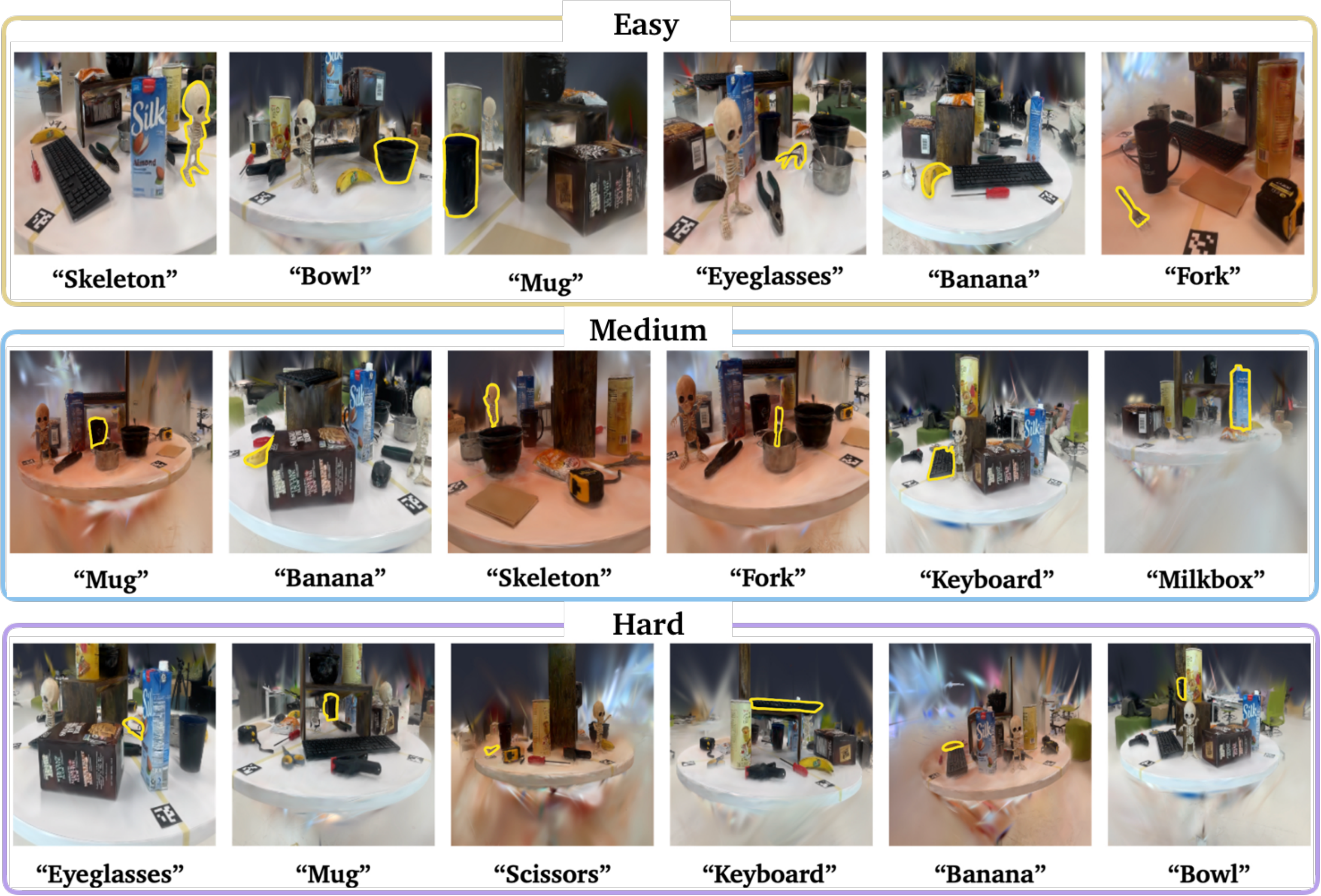}
    \caption{\textbf{GS-Random initialization difficulty.} Initial observations and target query for different initial-pose difficulty levels in the GS-Random scene.
    }
    \label{fig:gs-init-exp}
\end{figure*}

To systematically evaluate our framework under various settings, we define task difficulty across two dimensions: \emph{scene difficulty}, which is determined by the diversity of objects present and the level of compactness, and \emph{initial-pose difficulty}, which is determined by a heuristics function that is computed from three factors: initial detection confidence, ground truth distance to target, and level of occlusion of the bounding box. 
The thresholds for these factors are environment-dependent to account for scene dimensions, detection qualities, etc.
Though the difficulty metrics are not strictly quantifiable, from empirical results we observe consistent trends in degrading performance as task difficulty increases, and show promising trend on WoMAP's robustness to exhibiting less performance degradation compared to the baselines. Figure \ref{fig:pb-init-exp} and \ref{fig:gs-init-exp} show a more comprehensive visualization of different initial condition levels in two PyBullet and Gaussian Splat environments to offer readers a better intuition.

\subsection{Hardware Experiment Setup}

We evaluate our policy trained entirely in the Gaussian Splat directly on a TidyBot platform with a Panda Franka arm. With a mobile base and a 6-DOF arm equipped with an onboard Intel RealSense camera, the TidyBot is well suited for active object localization tasks since it has a larger effective workspace, compared to tabletop, fixed-base Franka robots that are constrained to mostly top-down views. We interface with the TidyBot through an onboard NUC, which publishes images to a desktop for inference using the world model or VLM.

\subsection{Evaluation Setup}
\label{ssec:appendix_baselines_ablations_metrics}

\subsubsection{Choice of Baselines and Ablations}

In the following sections, we discuss in more detail how we setup the evaluation framework, and motivation for the choice of baselines and ablations that we choose to include. Selecting the proper baseline is particularly challenging in our case. For one, many policies are not open-vocabulary and require more constrained problem/action spaces. Secondly, the setup of our task (large environment variations and observations from only an onboard camera) also makes finetuning state-of-the-art manipulation policies difficult, since they are not designed for the task. 
In our experiments, we compare WoMAP to a VLM-based planner with GPT-4o \cite{hurst2024gpt} as the VLM. We compute the gradients during planning using automatic differentiation in the WM-Random, WM-HR, and WoMAP.

\begin{figure*}
    \centering
    \ifbool{rss_format}
    {
    \newcommand{\figwidth}{0.9}
    }
    {
    \newcommand{\figwidth}{1.0}
    }
    \includegraphics[width=\figwidth\linewidth]{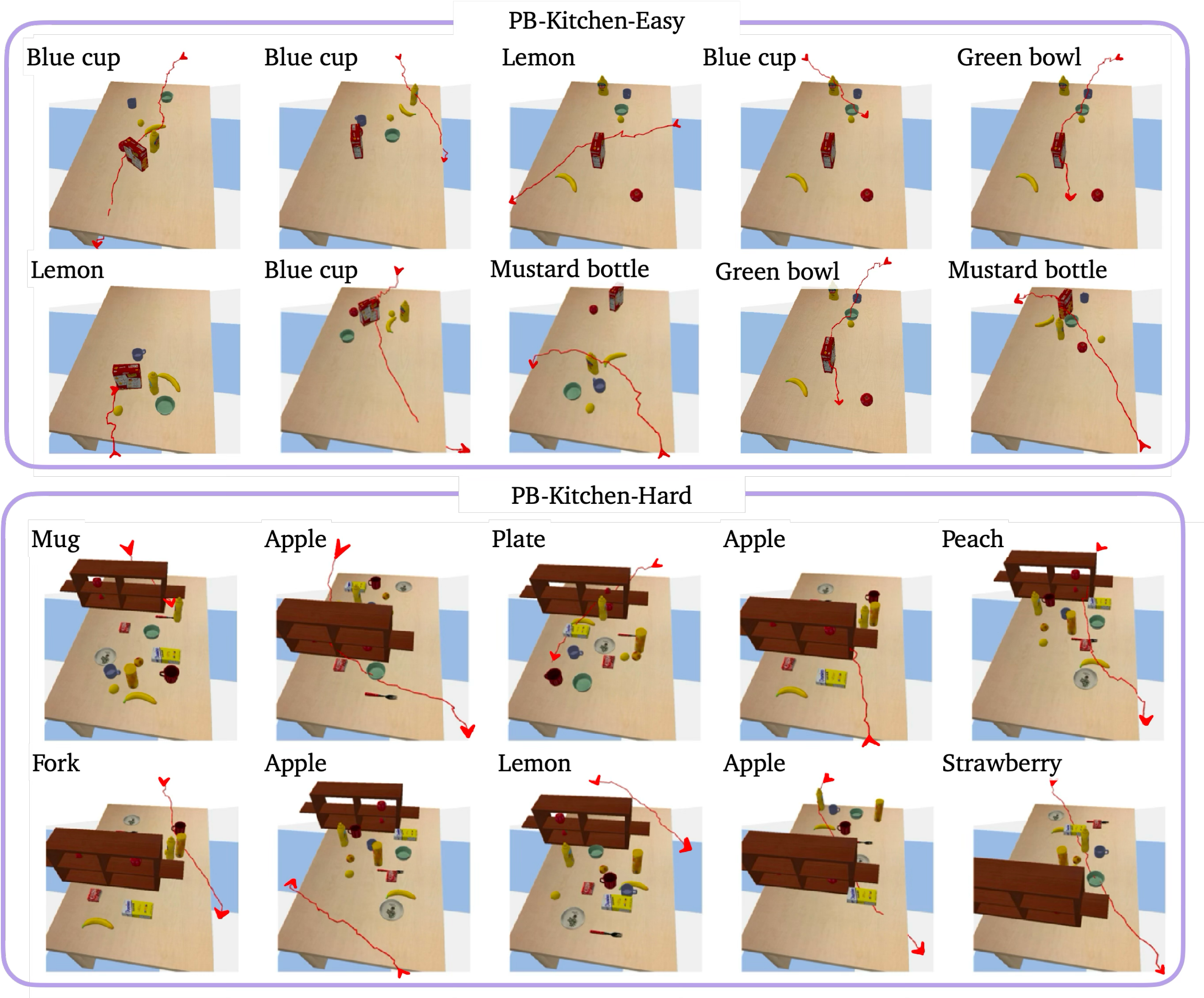}
    \caption{\textbf{Diffusion policy trajectory visualizations.} Top two rows: PB-Kitchen-Easy with easy initial conditions, Bottom two rows: PB-Kitchen-Hard with hard initial conditions.
    The DP generally moved forward in the direction the robot was initialized at, without showing any intelligent exploration behavior, e.g., looking behind objects.}
    \label{fig:dp-traj}
\end{figure*}

\textbf{Diffusion Policy.} 
One might expect the diffusion policy to perform better in our experiments.
Upon further investigation,
we observed that a \emph{single-task} diffusion policy (trained to localize a single object in an environment) performs well; however, the \emph{multi-task} diffusion policy (DP), which is more relevant to the active object localization problem, failed to perform well.
We found that the DP generally moved forward in the direction the robot was initialized at, without exhibiting any intelligent exploration behavior, e.g., looking behind objects and in occluded areas.
We visualize some of the DP trajectories in \Cref{fig:dp-traj} in the PB-Kitchen-Easy and PB-Kitchen-Hard scenes. The arrows in the figure indicate the direction of travel. From \Cref{fig:dp-traj}, we see that the DP does not seem to make intentional decisions to move towards the target objects, annotated in the figure.

\subsubsection{Choice of Metrics}

We make two key observations when designing the evaluation metrics. First, in order to make sure that the final observation is of good quality, our \emph{success score} is defined to be $1$ if both (i) the detection confidence labeled by GroundingDINO is above a certain threshold, and (ii) the labeled bounding box proportion is above a certain threshold.
Since GroundingDINO's detection confidence and bounding box size on different objects can behave differently depending on the object identity, shape, or size, we choose object-specific scaling parameters from the best view.

Secondly, the quality of task completion should also be dependent on the amount of distance traveled to reach the target as some policies are more efficient than others. To this end, we define the \emph{efficiency score} as ${\text{efficiency} = r * \exp(- d / d^{\star})}$, where $r$ denotes the success rate, $d$ denotes the distance traveled, and $d^{\star}$ the optimal distance to the object. In practice, we use an estimate of $d^{\star}$.

\subsection{Additional Experiment Results}

\subsubsection{Ablation Studies on Training Data}
\label{appendix:training-data-ablation}
Though we performed all of our experiments with a fixed training setup (50 scenes-500 trajectories for the PyBullet environments and 10 scenes-300 trajectories for the Gaussian Splat/Real environments), we conducted additional ablation studies to investigate the influence of the size and diversity of the training data on model performance. 
We perform this ablation study in the PB-Kitchen-Easy scene, since we can easily vary the size and diversity of the training data in PyBullet.

\begin{figure*}
    \centering
    \includegraphics[width=0.7\linewidth]{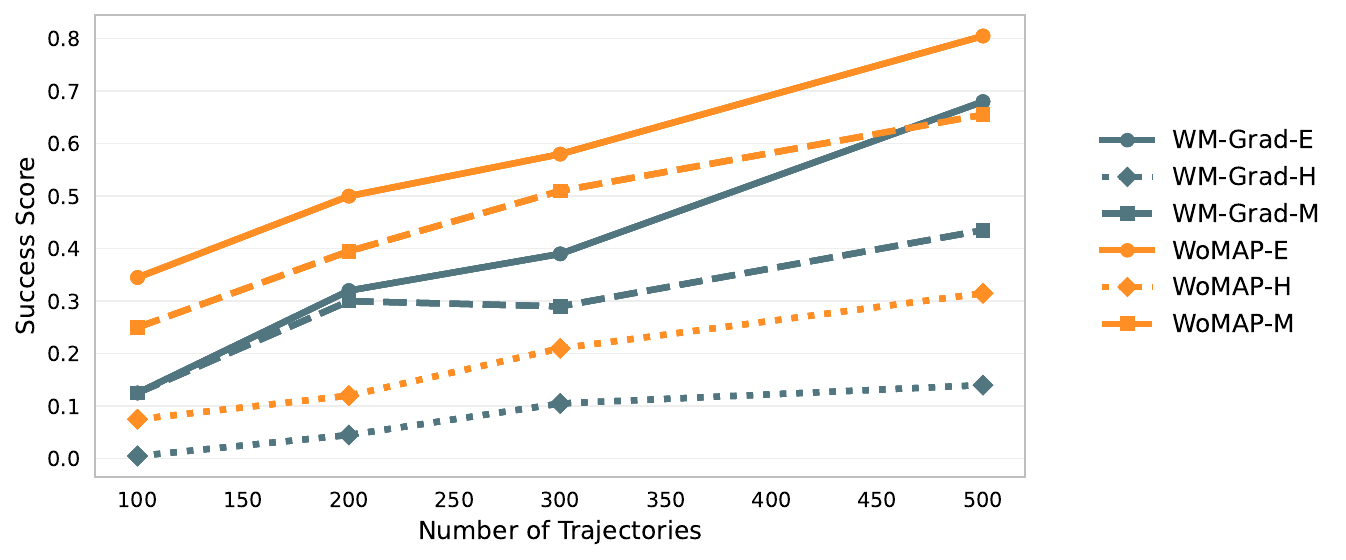}
    \caption{Average success score for different number of trajectories sampled from 50 scenes in the PB-Kitchen-Easy task. We observe a positive correlation between the number of training trajectories and the success rate of WM-Grad and WoMAP across different initial conditions: Easy (E), Medium (M), and Hard (H).
    With only $200$ training trajectories, WoMAP outperforms the VLM and DP baselines (see \Cref{fig:baselines_pybullet_plot_dp_binary_success}).}
    \label{fig:numtraj-trend}
\end{figure*}

In Figure~\ref{fig:numtraj-trend}, we observe that the performance of WoMAP and WM-Grad increases as the total number of trajectories in the training dataset increases.
We show the performance of WM-Grad alongside WoMAP to indicate the base performance of the world model without VLM action proposals.
As expected, WoMAP's performance is strongly correlated with the size of the training data, which influences the dynamics and rewards prediction accuracy of the world model.
Notably, even with only $200$ training trajectories, WoMAP outperforms the VLM and DP baselines by about $1033\%$ and $17\%$, respectively, (see \Cref{fig:baselines_pybullet_plot_dp_binary_success}), showing its remarkable data efficiency.

\begin{figure*}
    \centering
    \includegraphics[width=\linewidth]{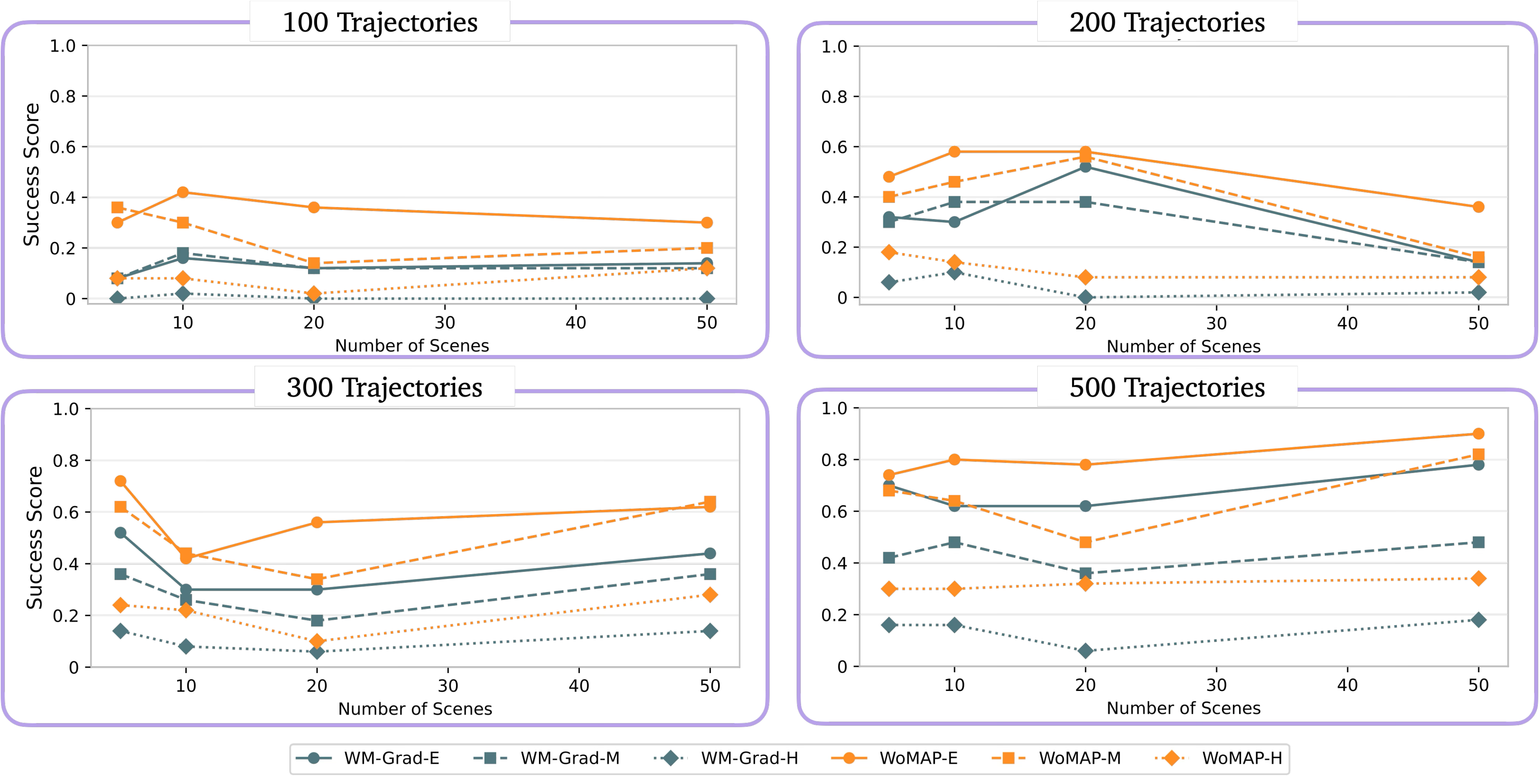}
    \caption{Average success score for total number of trajectories $\in \{100, 200, 300, 400\}$ sampled from different numbers of scenes in the PB-Kitchen-Easy task. When trained with very few trajectories (e.g., $100$ or $200$ trajectories), the model performance decreases as the number of training scenes increases, due to the tradeoff between learning scene-specific features versus generalizable features. However, we observe a positive correlation between the number of training scenes and the success rate when training with more trajectories, e.g., $500$ trajectories.}
    \label{fig:numscenes-trend}
\end{figure*}

Figure~\ref{fig:numscenes-trend} shows the performance of WoMAP as we vary the scene diversity while keeping the total number of trajectories fixed. 
Overall, we do not observe a strong correlation between the number of training scenes and the success score. However, when training with fewer trajectories (e.g., $100$ or $200$ trajectories), the performance of the models decreases with the number of scenes. This finding may be attributed to the tradeoff between learning more specialized (scene-specific) features versus more generalizable features across multiple scenes, with a tight budget on the number of training trajectories.
In contrast, when training with $500$ trajectories, we find that the success rate improves as the number of scene increases, resulting in much higher success rates.

\subsubsection{Full Experiment Results}

We provide the full results for the experiments used to compute the values in Figure \ref{fig:baselines_pybullet_plot_dp_binary_success} and \ref{fig:baselines_gsplat_plot_dp_binary_success} in Table~\ref{tab:full-exp-results}. Each task under a given environment and initialization difficulty is performed under 50 independent trials with randomized scene configuration and initial position.

\begin{table*}[th]
    \centering
    \caption{Full Experiment Results.}
    \label{tab:full-exp-results}
\begin{tabular}{lcccccccc}
\toprule
IC: Easy & \multicolumn{2}{c}{\textbf{PB-Office}} & \multicolumn{2}{c}{\textbf{PB-Kitchen-Easy}} & \multicolumn{2}{c}{\textbf{PB-Kitchen-Medium}} & \multicolumn{2}{c}{\textbf{PB-Kitchen-Hard}} \\
\cmidrule(lr){2-3} \cmidrule(lr){4-5}\cmidrule(lr){6-7} \cmidrule(lr){8-9}
\textbf{Planner}  & Success & Efficiency & Success & Efficiency & Success & Efficiency & Success & Efficiency \\
\midrule
VLM & 0.30 & 0.28 & 0.06 & 0.05 & 0.00 & 0.00 & 0.02 & 0.02 \\
DP & 0.28 & 0.25 & 0.28 & 0.25 & 0.22 & 0.19 & 0.16 & 0.15 \\
WM-CEM & 0.30 & 0.24 & 0.18 & 0.13 & 0.10 & 0.08 & 0.16 & 0.13 \\
WM-Grad & 0.60 & 0.41 & 0.78 & 0.50 & 0.66 & 0.41 & 0.66 & 0.46 \\
WM-HR & 0.74 & 0.63 & 0.74 & 0.61 & 0.68 & 0.56 & 0.62 & 0.54 \\
WoMAP & 0.88 & 0.73 & 0.90 & 0.72 & 0.78 & 0.60 & 0.80 & 0.68 \\

\bottomrule
\end{tabular}
\vspace{5pt}
\begin{tabular}{lcccccccc}
\toprule
IC: Medium & \multicolumn{2}{c}{\textbf{PB-Office}} & \multicolumn{2}{c}{\textbf{PB-Kitchen-Easy}} & \multicolumn{2}{c}{\textbf{PB-Kitchen-Medium}} & \multicolumn{2}{c}{\textbf{PB-Kitchen-Hard}} \\
\cmidrule(lr){2-3} \cmidrule(lr){4-5}\cmidrule(lr){6-7} \cmidrule(lr){8-9}
\textbf{Planner}  & Success & Efficiency & Success & Efficiency & Success & Efficiency & Success & Efficiency \\
\midrule
VLM & 0.24 & 0.22 & 0.04 & 0.04 & 0.04 & 0.04 & 0.02 & 0.02 \\
DP & 0.42 & 0.36 & 0.26 & 0.23 & 0.28 & 0.25 & 0.08 & 0.07 \\
WM-CEM & 0.26 & 0.22 & 0.24 & 0.17 & 0.24 & 0.18 & 0.16 & 0.11 \\
WM-Grad & 0.58 & 0.39 & 0.48 & 0.30 & 0.52 & 0.34 & 0.46 & 0.32 \\
WM-HR & 0.74 & 0.61 & 0.66 & 0.52 & 0.50 & 0.41 & 0.44 & 0.38 \\
WoMAP & 0.70 & 0.55 & 0.82 & 0.63 & 0.64 & 0.52 & 0.50 & 0.41 \\
\bottomrule
\end{tabular}
\vspace{5pt}
\begin{tabular}{lcccccccc}
\toprule
IC: Hard & \multicolumn{2}{c}{\textbf{PB-Office}} & \multicolumn{2}{c}{\textbf{PB-Kitchen-Easy}} & \multicolumn{2}{c}{\textbf{PB-Kitchen-Medium}} & \multicolumn{2}{c}{\textbf{PB-Kitchen-Hard}} \\
\cmidrule(lr){2-3} \cmidrule(lr){4-5}\cmidrule(lr){6-7} \cmidrule(lr){8-9}
\textbf{Planner}  & Success & Efficiency & Success & Efficiency & Success & Efficiency & Success & Efficiency \\
\midrule
VLM & 0.08 & 0.05 & 0.00 & 0.00 & 0.00 & 0.00 & 0.00 & 0.00 \\
DP & 0.14 & 0.10 & 0.32 & 0.28 & 0.20 & 0.18 & 0.18 & 0.16 \\
WM-CEM & 0.06 & 0.04 & 0.02 & 0.01 & 0.02 & 0.02 & 0.02 & 0.02 \\
WM-Grad & 0.24 & 0.11 & 0.18 & 0.11 & 0.16 & 0.09 & 0.16 & 0.09 \\
WM-HR & 0.38 & 0.26 & 0.24 & 0.19 & 0.24 & 0.19 & 0.22 & 0.19 \\
WoMAP & 0.58 & 0.39 & 0.34 & 0.25 & 0.40 & 0.32 & 0.34 & 0.28 \\
\bottomrule
\end{tabular}
\vspace{5pt}
\begin{tabular}{lcccccc}
\toprule
IC: Easy & \multicolumn{2}{c}{\textbf{GS-Kitchen}} & \multicolumn{2}{c}{\textbf{GS-Office}} & \multicolumn{2}{c}{\textbf{GS-Random}} \\
\cmidrule(lr){2-3} \cmidrule(lr){4-5}\cmidrule(lr){6-7} 
\cmidrule(lr){2-3} \cmidrule(lr){4-5}\cmidrule(lr){6-7}
\textbf{Planner}  & Success & Efficiency & Success & Efficiency & Success & Efficiency \\
\midrule
VLM & 0.12 & 0.11 & 0.14 & 0.13 & 0.08 & 0.07 \\
DP & 0.50 & 0.47 & 0.54 & 0.49 & 0.28 & 0.26 \\
WM-CEM & 0.58 & 0.48 & 0.40 & 0.35 & 0.40 & 0.30 \\
WM-Grad & 0.84 & 0.67 & 0.86 & 0.63 & 0.76 & 0.56 \\
WM-HR & 0.78 & 0.66 & 0.96 & 0.83 & 0.58 & 0.49 \\
WoMAP & 0.86 & 0.74 & 0.96 & 0.77 & 0.72 & 0.59 \\
\bottomrule
\end{tabular}

\begin{tabular}{lcccccc}
\toprule
IC: Medium & \multicolumn{2}{c}{\textbf{GS-Kitchen}} & \multicolumn{2}{c}{\textbf{GS-Office}} & \multicolumn{2}{c}{\textbf{GS-Random}} \\
\cmidrule(lr){2-3} \cmidrule(lr){4-5}\cmidrule(lr){6-7}
\textbf{Planner}  & Success & Efficiency & Success & Efficiency & Success & Efficiency \\
\midrule
VLM& 0.06 & 0.06 & 0.06 & 0.05 & 0.02 & 0.01 \\
DP & 0.42 & 0.36 & 0.32 & 0.29 & 0.12 & 0.11 \\
WM-CEM & 0.26 & 0.20 & 0.22 & 0.18 & 0.20 & 0.13 \\
WM-Grad & 0.50 & 0.32 & 0.50 & 0.30 & 0.36 & 0.23 \\
WM-HR & 0.56 & 0.45 & 0.66 & 0.56 & 0.30 & 0.23 \\
WoMAP & 0.68 & 0.55 & 0.72 & 0.55 & 0.46 & 0.36 \\
\bottomrule
\end{tabular}

\begin{tabular}{lcccccc}
\toprule
IC: Hard & \multicolumn{2}{c}{\textbf{GS-Kitchen}} & \multicolumn{2}{c}{\textbf{GS-Office}} & \multicolumn{2}{c}{\textbf{GS-Random}} \\
\cmidrule(lr){2-3} \cmidrule(lr){4-5}\cmidrule(lr){6-7}
\textbf{Planner}  & Success & Efficiency & Success & Efficiency & Success & Efficiency \\
\midrule
VLM & 0.00 & 0.00 & 0.02 & 0.02 & 0.00 & 0.00 \\
DP & 0.18 & 0.16 & 0.14 & 0.11 & 0.08 & 0.06 \\
WM-CEM & 0.08 & 0.06 & 0.08 & 0.05 & 0.02 & 0.01 \\
WM-Grad & 0.14 & 0.09 & 0.22 & 0.11 & 0.00 & 0.00 \\
WM-HR & 0.18 & 0.15 & 0.32 & 0.25 & 0.08 & 0.07 \\
WoMAP & 0.24 & 0.18 & 0.48 & 0.34 & 0.12 & 0.08 \\
\bottomrule
\end{tabular}
    \vspace{3pt}
\end{table*}

\subsubsection{Semantic Generalization Experiment Results}
\label{appendix-subsec:semantic-generalization}
As discussed in \Cref{ssec:exp_generalization_to_novel_task_conditions}, we evaluate WoMAP's semantic zero-shot generalization to novel tasks.
We quantify the semantic similarity between different tasks using the cosine similarity between language embeddings computed from the task instructions by Sentence-Bert~\cite{reimers-2019-sentence-bert}. 
For seen target objects, we vary the task instructions to capture the diversity of possible user descriptions, e.g., asking WoMAP to find a ``sweet thing" or a ``yellow fruit" with the goal of locating a banana. 
We utilize partial success scores to better evaluate the degradation in performance with more semantically dissimilar target object queries, where the success scores is linearly scaled based on the bounding-box proportion and the detection confidence. 
From \Cref{fig:generalization-bsm}, we find that WoMAP achieves strong generalization with a correlation coefficient between $0.6$ and $0.71$ across the three object categories. Likewise in \Cref{fig:generalization-bsm}, we observe that WoMAP generalizes well to unseen target objects. For example, WoMAP localizes a pair of ``pliers" and a ``hammer," which WoMAP was not trained on.
We attribute the strong generalization performance of WoMAP to the generalizable semantic features captured in WoMAP's latent space.

\subsubsection{Additional Discussion}

Here, we provide additional discussion for the main results in Section~\ref{sec:experiments}.

\textbf{World Model-Heuristics (WM-HR) Baseline}:
We use the set of candidate actions provided to the VLM as the heuristic actions in the WM-HR baseline. This clever set of actions enabled the WM-HR to perform similarly to WoMAP, since the world model performed well at evaluating these candidate actions and selecting the most promising one. However, the WM-HR baseline lacks any high-level intelligence, posing a limitation in practical situations, since a brute-force approach would not scale, in general.

\textbf{Sim-to-Real Performance for VLM}: We observe that the performance of the VLM drops significantly when moving from simulation to the real-world. This drop in performance is in part due to the limitations in directly executing the VLM's action proposals on the physical robot. We find that the robot reaches its range limits often when executing the VLM action proposals, which effectively ends the experiments, leading to an increase in the failure of the VLM. Further, the VLM occasionally struggled to fully approach the object in the real-world and often stopped a good distance away from the object.

\subsubsection{Extension: Non-tabletop Scenes}

Despite using tabletop scenes for straightforward, standardized benchmarking in this paper, we also illustrate that our framework works for more general environments. We provide a video in the supplementary material, demonstrating WoMAP on a larger scale, particularly in a living room.